\makeatletter
\let\my@xfloat\@xfloat
\makeatother

\documentclass[%
	final,
	notitlepage,
	narroweqnarray,
	inline,
	twoside,
	]{ieee}
	
\makeatletter
\def\@xfloat#1[#2]{
	\my@xfloat#1[#2]%
	\def\baselinestretch{1}%
	\@normalsize \normalsize
}
\makeatother

\usepackage{color}
\usepackage{soul}
\usepackage{graphicx}
\usepackage{amsfonts}

\usepackage{amsmath}
\usepackage{cases}
\usepackage{amssymb}
\usepackage{mathtools}
\usepackage{leftidx}
\usepackage{ragged2e}
\usepackage{multirow}
\newtheorem{lemma}{Lemma}
\newcommand{\argmin}{\operatornamewithlimits{argmin}}
\newcommand*{\QEDA}{\hfill\ensuremath{\blacksquare}}
\newcommand{\vtxt}[1]{\textcolor{red}}

\begin{document}
\title[A NOVEL METHOD FOR EXTRINSIC CALIBRATION OF A 2D LRF AND A CAMERA]{A Novel Method for the Extrinsic Calibration of \\ a 2D Laser Rangefinder and a Camera}

\author[DONG AND ISLER]{
\thanks{A preliminary version of this work was presented in~\protect\cite{Dong}. This extended version includes spatial-ambiguity explanation, the complete proof of uniqueness, the analytical solution, as well as calibration details and additional experiments.}
\thanks{This work is supported in part by NSF Award 1317788, USDA Award MIN-98-G02 and the MnDrive initiative.}
Wenbo Dong\member{Student 
       Member}, and Volkan Isler\member{Senior Member}\authorinfo{W. Dong and V. Isler are with the Department of Computer Science and Engineering, University of Minnesota, Twin Cities, MN, 55455, USA. E-mail: \{dong, isler\}@cs.umn.edu}}

\journal{}
\ieeecopyright{????--????/??\$??.?? \copyright\ ???? IEEE}
\lognumber{xxxxxxx}
\pubitemident{S ????--????(??)?????--?}
\loginfo{Manuscript received ??? ??, ????.}

\confplacedate{????, ???, ??? ??--??, ????}
\maketitle
\begin{abstract} 
We present a novel method for extrinsically calibrating a camera and a 2D Laser Rangefinder (LRF) whose beams are invisible from the camera image. We show that point-to-plane constraints from a single observation of a V-shaped calibration pattern composed of two non-coplanar triangles suffice to uniquely constrain the relative pose between two sensors. Next, we present an approach to obtain analytical solutions using point-to-plane constraints from single or multiple observations. Along the way, we also show that previous solutions, in contrast to our method, have inherent ambiguities and therefore must rely on a good initial estimate. Real and synthetic experiments validate our method and show that it achieves better accuracy than previous methods.
\end{abstract}

\section{Introduction}\label{sec:introduction}
\PARstart Many robotics systems rely on cameras and laser range finders to compute environment geometry~\cite{Douillard}. Two dimensional (2D) Laser Range Finders (LRFs) which measure depth along a single plane are commonly used due to their low weight, low cost and low power requirements.

Taking advantage of measurements from an LRF or a LIDAR combined with a camera, however, requires precise knowledge of the relative pose (orientation and position) between them. This is a classical extrinsic calibration problem where the objective is to determine the transformation between two coordinate frames. 
Establishing correspondences between two sensors is easier for 3D LIDARs since distinct features can be identified both among laser points and in the camera image.
Existing methods include 3D LIDAR-camera calibration by using a circle-based calibration target~\cite{circleBased} and an arbitrary trihedron~\cite{trihedron}.

Extrinsic calibration of a 2D LRF is more challenging
because a 2D LRF produces only a single scanning plane for each pose which is invisible from the regular camera. This makes it difficult to find correspondences.
Therefore, additional constraints must be used.
\emph{We note that if we were given the correspondences between laser points and their images (e.g. IR camera) the extrinsic calibration problem reduces to a standard P$n$P (Perspective-n-Point) computation~\cite{PNP}. However, in our case, these correspondences are unknown.}

There is a large body of work on the LRF-camera calibration. One of the earliest methods is presented by Zhang and Pless~\cite{Zhang} using points-on-plane constraints. However, only two degrees of freedom are constrained for the relative pose between the camera and the LRF from a single input observation. Therefore, this method requires a large number of different observations with a wide range of views (more than 20 snapshots) for accuracy. Vasconcelos et al.~{\cite{Vasconcelos}} presented a minimal solution by forming a  perspective-three-point (P3P) problem to address disadvantages in~{\cite{Zhang}}. Zhou~{\cite{Zhou}} further proposed an algebraic method for extrinsic calibration. Both techniques require multiple observations (at least three), and have inherent degeneracy where intersecting lines of two planes with the third plane are parallel, and three intersecting points of laser segments from three input observations are on a danger cylinder.

\begin{figure}[t]
	\centering
	\includegraphics[width=0.6\columnwidth]{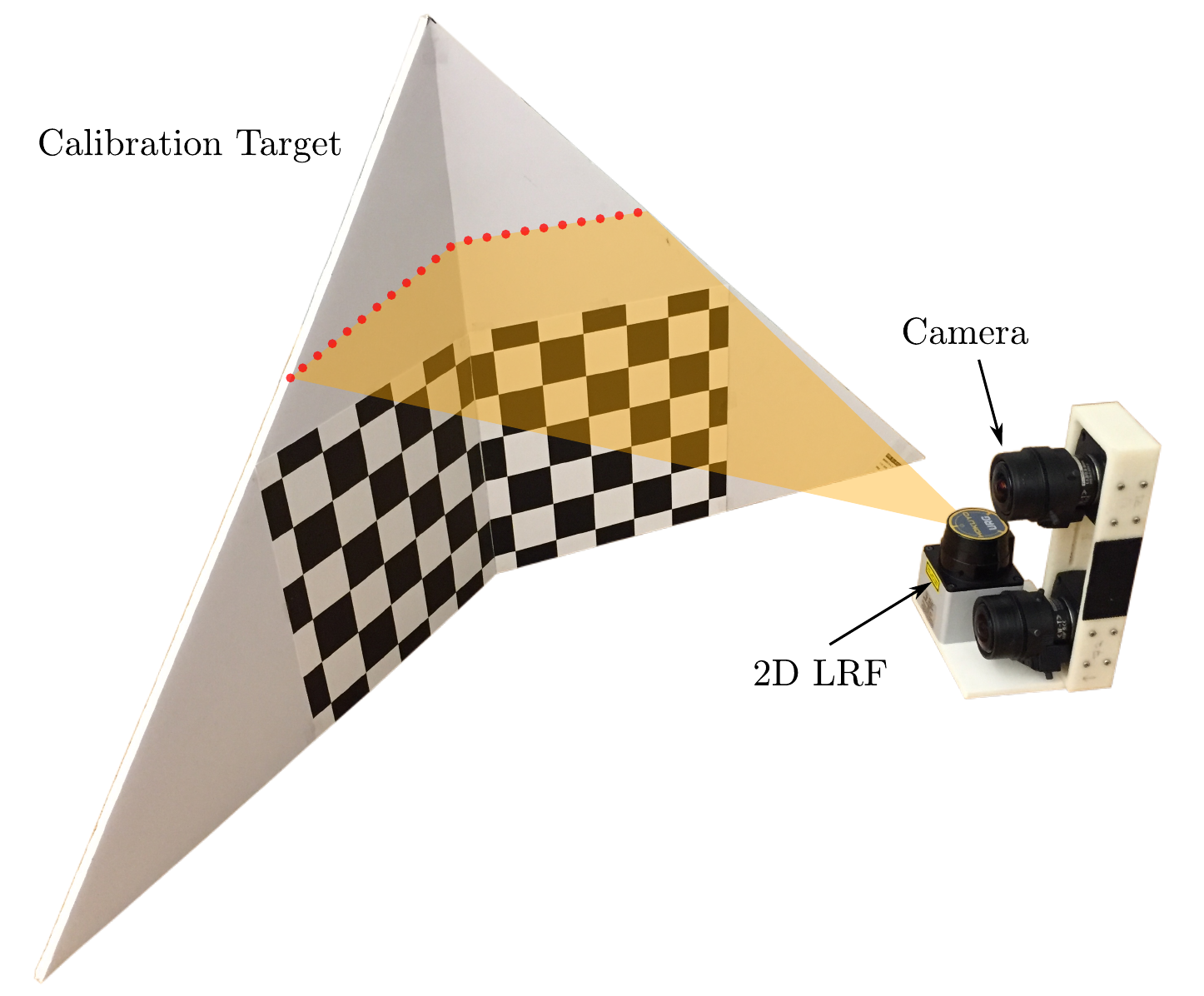}
	\caption{The calibration system incorporating a calibration target and a capture rig; Left: The calibration target formed by two triangular boards with a checkerboard on each triangle; Right: The capture rig consisting of a 2D LRF and stereo cameras. (Only one camera is involved in the calibration problem, the other is just for testing in real experiment.)}
	\label{fig:cameralaserRig}
	\vspace*{-2mm}
\end{figure}

\begin{figure*}[t]
	\centering
	\includegraphics[width=0.95\textwidth]{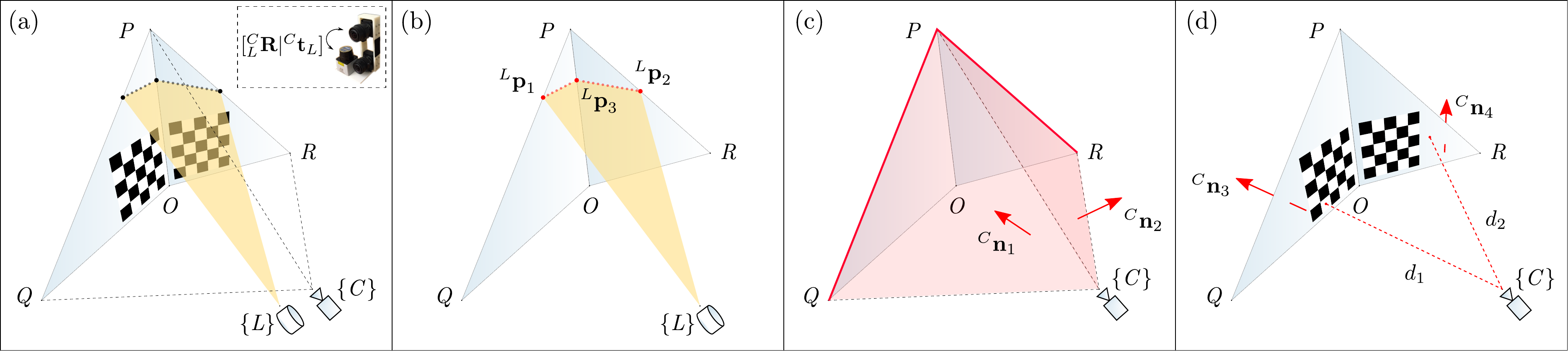}
	\caption{A single pair of LRF-camera observations of our calibration target with the definition of parameters for geometry constraints. (a): The output of the 2D LRF-camera calibration is the relative transformation ${_L^C}\mathbf{R}$ and ${^C}\mathbf{t}{_L}$. (b): The input data from the LRF are three laser points ${^L}\mathbf{p}_1$, ${^L}\mathbf{p}_2$ and ${^L}\mathbf{p}_3$. (c): Two input normals from the camera are ${^C}\mathbf{n}_1$ and ${^C}\mathbf{n}_2$. (d): The other input data from the camera are two normals ${^C}\mathbf{n}_3$ and ${^C}\mathbf{n}_4$ and two distances $d_1$ and $d_2$.}
	\label{fig:calibTarget}
	\vspace*{-1mm}
\end{figure*}

With point-on-line constraints, the approaches in~\cite{Li} and ~\cite{Wasielewski} use a black triangular board and a V-shaped calibration target respectively. The results from these two methods are not accurate due to the sparse sampling of laser points. Further, a large number (usually more than 100) of images are needed to compensate for the lack of constraints for each input observation. Based on the ideas in~\cite{Li} and~\cite{Wasielewski} (minimizing the projection distance on the image between intersected laser points and the feature lines), the authors in~\cite{Kwak} also propose to use a V-shaped calibration target. They increase the laser points' sampling for each observation by introducing more feature lines and virtual end-points, but the same drawback still exists as in~\cite{Wasielewski} and~\cite{Kwak}. Therefore, they still need a large amount (around 50) of different observations to achieve a reasonable result.

The method in~\cite{Naroditsky} provides an analytical solution using a white board with a black band in the middle. It needs only six different observations. Similarly, the authors in~\cite{Guo} give an analytical solution to this problem using a white board with a black line in the middle. Compared with~\cite{Naroditsky}, it further computes the optimal least-squares solution to improve the robustness to noise. The analytical solutions in~{\cite{Naroditsky}} and~{\cite{Guo}} are obtained by minimizing the sum of points-on-plane errors, where only perspective planes from the image are considered instead of general 3D planes. However, both of these two methods still cannot avoid using a large number of different observations for accuracy because of the insufficient constraints for each input observation.

The work described in~\cite{Ruben}, presents an approach which only requires the observation of a scene corner (orthogonal trihedron) commonly found in human-made environments. This method builds line-to-plane and point-to-plane constraints, which requires at least two input observations. However, the calibration accuracy highly depends right angles between three orthogonal planes, which are difficult to be made exactly $90^\circ$ in practice.
When multiple observations from different views are needed for additional accuracy, the right angle between 
two walls often affects the laser measurement: scanned laser line on one wall is curved if laser beams point almost perpendicular towards the other wall for a good view. Our method accommodates an arbitrary obtuse angle in our calibration target (See Fig.~\ref{fig:cameralaserRig}) so that it can adjust the view angle between the pattern and linear beams.

The authors in~\cite{Hu} further extend the work~\cite{Ruben} by deriving a minimal solution from a single input observation. The solution, however, is obtained by two procedures (calibration between the trihedron and the LRF, and calibration between the trihedron and the camera), and thus has accumulated error due to the data noise in each procedure. Specifically, in calibration between the trihedron and the camera, they determine the scale of the translation by using the actual length of two edges of the trihedron which is inconvenient to be built and difficult to be measured accurately.

In theory, LRF-camera calibration from a single input observation is important since it means that the geometric constraints from a single view is sufficient. In practice, it further implies that users, when taking  multiple input observations for further accuracy, do not need to be concerned about degenerate cases in which the input observation is invalid.
Our triangular V-shaped calibration target (See Fig.~\ref{fig:cameralaserRig}) has two checkerboards, which are simultaneously and accurately estimated in camera calibration. Further, the angle between two triangular boards of the target can be arbitrary which makes it convenient to build. We study this extrinsic calibration problem and make the following contributions:
\begin{itemize}
\item We show that by minimizing the sum of points-on-plane errors, a single observation of two non-coplanar triangles sharing a common side (See Fig.~\ref{fig:cameralaserRig}) suffices to unambiguously solve the calibration problem.
\item Even though planar, triangular or V-shaped rectangular patterns have already been proposed to solve the calibration problem, we show that previous methods do not sufficiently constrain the calibration problem to allow for a unique solution. Therefore, they rely on a large number of measurements and a good initial estimate.
\item We also present a robust analytical solution to the system of points-on-plane constraints for calibration from a single observation in the presence of noise.
\item For additional accuracy, we show that by using only a few additional observations, our method achieves significantly smaller error than existing methods.
\end{itemize}

\section{Spatial Ambiguities In Previous Methods} \label{relatedWork}
The objective of 2D LRF-camera calibration is to obtain the relative pose between these two sensors: the orientation ${_L^C}\mathbf{R}$ and position ${^C}\mathbf{t}{_L}$ of the LRF frame $\{ L\}$ with respect to (w.r.t.) the camera frame $\{ C\}$ (See Fig.~\ref{fig:calibTarget}). Spatial ambiguity means that there are infinite solutions for ${_L^C}\mathbf{R}$ and ${^C}\mathbf{t}{_L}$ from a single input observation of the calibration target.

Without loss of generality, the laser scanning plane is defined as the plane $Y_L = 0$ such that we do not have an explicit dependence on the second column vector $\mathbf{r}_2$ of ${_L^C}\mathbf{R}$ when a 3D laser point ${^L}\mathbf{p} = [x_L, 0, z_L]^{\top}$ is transformed to the point ${^C}\mathbf{p} = [x_C, y_C, z_C]^{\top}$ in the camera frame by ${^C}\mathbf{p} = {_L^C}\mathbf{R}\cdot{^L}\mathbf{p} + {^C}\mathbf{t}_L$. Since ${_L^C}\mathbf{R}$ is an orthonormal matrix, we have three constraints for its first and third columns ($\mathbf{r}_1$ and $\mathbf{r}_3$)
\begin{equation} \label{constraint4}
\mathbf{r}_3^{\top} \mathbf{r}_3 = 1, \quad \mathbf{r}_1^{\top} \mathbf{r}_1 = 1, \quad \mathbf{r}_3^{\top} \mathbf{r}_1 = 0 .
\end{equation}
Thus, we need at least six additional geometry constraints for solving nine unknowns (in $\mathbf{r}_1$, $\mathbf{r}_3$ and ${^C}\mathbf{t}{_L}$).

\begin{figure*}[t]
	\centering
	\includegraphics[width=0.95\textwidth]{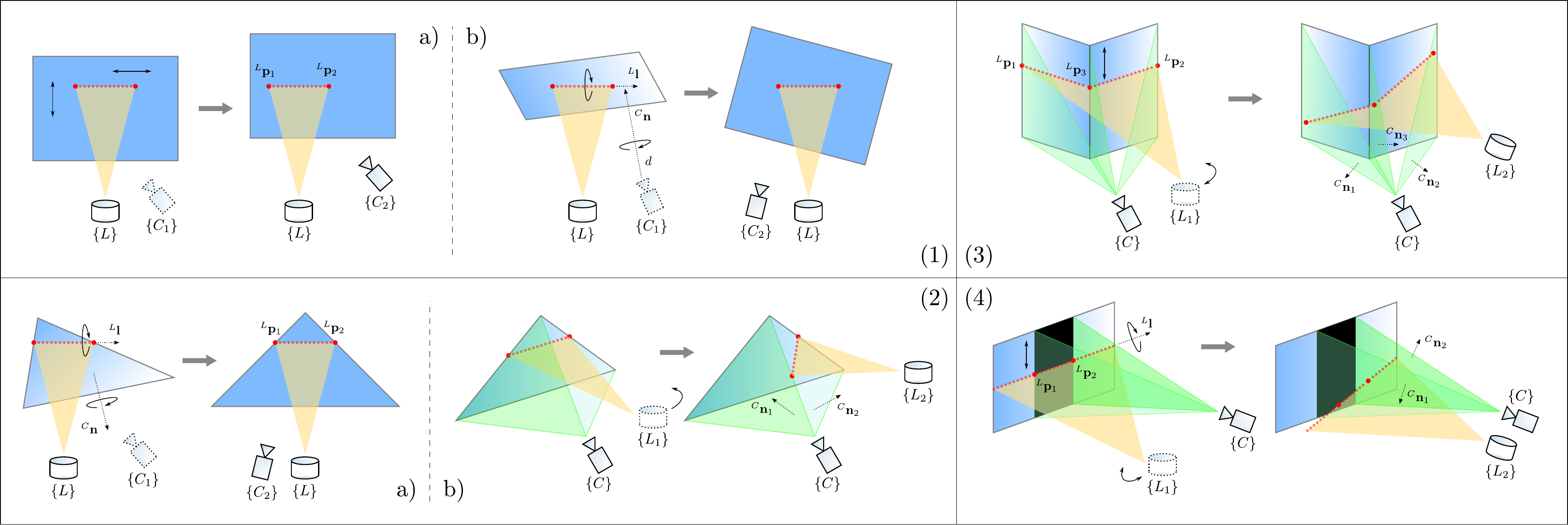}
	\caption{Existing methods do not sufficiently constrain the problem from a single input observation. (1): In the approach of Zhang and Pless~\protect\cite{Zhang}, essentially, only two laser points ${^L}\mathbf{p}_1$ and ${^L}\mathbf{p}_2$ are constrained on the calibration board, where $d$ is the distance from the camera to the board. a): The camera frame $C_1$ with the board can be moved horizontally and vertically along the board to $C_2$. b): The camera frame $C_1$ with the board are first rotated along the board normal ${^C}\mathbf{n}$ and then rotated along the laser line direction ${^L}\mathbf{l}$ to reach $C_2$.
	(2): In the approach of Li et al.~\protect\cite{Li}, only two geometry constraints are obtained. a): The camera frame $C_1$ with the triangular calibration board are first rotated long the board normal ${^C}\mathbf{n}$ and then rotated along the laser line direction ${^L}\mathbf{l}$ to reach $C_2$ such that ${^L}\mathbf{p}_1$ and ${^L}\mathbf{p}_2$ are still on the border lines. b): The LRF frame $L_1$ can be moved and rotated to get $L_2$ as long as ${^L}\mathbf{p}_1$ and ${^L}\mathbf{p}_2$ are on green 3D perspective planes of 2D detected board edges.
	(3): In methods~\protect\cite{Wasielewski} and~\protect\cite{Kwak}, three geometry constraints are ${^C}\mathbf{n}_i^{\top} \left({_L^C}\mathbf{R} \cdot {^L}\mathbf{p}_i + {^C}\mathbf{t}{_L}\right) = 0$ where ${^C}\mathbf{n}_i$ with $i = 1, 2, 3$ are three normals of 3D perspective planes.
To reach $L_2$, the LRF frame $L_1$ with the laser line segments can be moved vertically along the calibration target and then rotated as long as ${^L}\mathbf{p}_1$, ${^L}\mathbf{p}_2$ and ${^L}\mathbf{p}_3$ lie on their green 3D perspective planes.
	(4): In approaches~\protect\cite{Guo} and~\protect\cite{Naroditsky}, geometry constraints (up to two) are ${^C}\mathbf{n}_i^{\top} \left({_L^C}\mathbf{R} \cdot {^L}\mathbf{p}_i + {^C}\mathbf{t}{_L}\right) = 0$ where ${^C}\mathbf{n}_i$ with $i = 1, 2$ are two normals of 3D perspective planes. Base on the same principle, the LRF frame $L_1$, in order to get $L_2$, can be moved vertically along the board, rotated along the laser line direction ${^L}\mathbf{l}$, and also rotated together with the line segment to make sure ${^L}\mathbf{p}_1$ and ${^L}\mathbf{p}_2$ still lie on corresponding perspective planes.}
	\label{fig:ambiguity}
	\vspace*{-2mm}
\end{figure*}

The spatial ambiguity is caused by the lack of sufficient geometric constraints from a single input observation. The disadvantage of insufficient constraints is that a large number of snapshots of the calibration target from different views are needed to reduce the ambiguity by minimizing the geometry cost function. A good initial estimate thus must be required otherwise the solution may converge to a local minimum which may not be the global minimum. However, this good initial estimate is not guaranteed in existing methods. Based on different type of calibration targets, previous methods can be classified into four categories: planar board with a checkerboard, triangle board, V-sharped target and rectangular board with a line or a band. Next, we detail the spatial ambiguity in each category.

\emph{Planar Board with a Checkerboard:} In the approach of Zhang and Pless~\protect\cite{Zhang}, all laser points must lie on the planar calibration pattern, described as ${^C}\mathbf{n}^{\top} \left({_L^C}\mathbf{R} \cdot {^L}\mathbf{p}_i + {^C}\mathbf{t}{_L}\right) = d$ in Fig.~\ref{fig:ambiguity}. Essentially, only two laser points are constrained from the single snapshot (two geometry constraints), and constraints of the rest of the laser points are redundant since they all belong to the same line segment. For the relative pose of the 2D LRF-camera pair, only two out of six degrees of freedom are constrained. The remaining four degrees have ambiguity such that there are infinite solutions for ${_L^C}\mathbf{R}$ and ${^C}\mathbf{t}{_L}$. As shown in Fig.~\ref{fig:ambiguity}, the calibration board together with the camera frame can be moved horizontally and vertically, and also can be rotated along two different axes without violating the geometry constraints.

\emph{Triangle Board:}
The work of Li et al.~\cite{Li} by using a triangular board does not improve the constraints in the method by Zhang and Pless~\cite{Zhang}: two laser end points ${^L}\mathbf{p}_1$ and ${^L}\mathbf{p}_2$ must lie on their corresponding border lines detected from the camera, represented as ${^C}\mathbf{n}_i^{\top} \left({_L^C}\mathbf{R} \cdot {^L}\mathbf{p}_i + {^C}\mathbf{t}{_L}\right) = 0$ where ${^C}\mathbf{n}_i$ with $i = 1, 2$ are the normals of the 3D perspective planes of 2D detected border lines. These two constraints remove the ambiguity of the horizontal translation and ``triangular plane'' removes the ambiguity of the vertical translation for ${^C}\mathbf{t}{_L}$. However, there are still three degrees of freedom that remain ambiguous for ${_L^C}\mathbf{R}$ (See Fig.~\ref{fig:ambiguity}). Essentially, the drawback is that the constraints are imposed on the 2D image: there exist uncertainties for a total of four unknown elements from views of depth and orientation (two linear geometry constraints plus three nonlinear constraints for ${_L^C}\mathbf{R}$ to solve nine unknowns). Additional details are explained in Section~\ref{uniqueSolution}.

\emph{V-sharped Target:}
The calibration target in~\cite{Wasielewski} and~\cite{Kwak} is formed as V-shaped by two rectangular boards. Three laser end points ${^L}\mathbf{p}_1$, ${^L}\mathbf{p}_2$ and ${^L}\mathbf{p}_3$ must lie on their corresponding board edges detected from the camera. Although the geometry constraints increase to three, the same drawbacks of spatial ambiguities still exist (the vertical translation of the calibration target as in~{\cite{Zhang}}, and the movement of laser points along their perspective planes as in~{\cite{Li}}). See Fig.~{\ref{fig:ambiguity}} for more details.

\emph{Rectangular Board with a Line or a Band:}
Methods in~\cite{Naroditsky} and~\cite{Guo} can be generalized as using a rectangular board with a black band (or a line) in the middle. Two laser end points ${^L}\mathbf{p}_1$ and ${^L}\mathbf{p}_2$ must lie on their band edges detected from the camera. With no more than two geometric constraints, they also suffer from spatial ambiguities. Thus, ${_L^C}\mathbf{R}$ and ${^C}\mathbf{t}{_L}$ have infinite solutions (See Fig.~\ref{fig:ambiguity}).

In contrast to previous methods, our method builds sufficient constraints, which guarantee the uniqueness of the solution for each input observation. In theory, we can use only one snapshot to calibrate the 2D LRF-camera rig. In practice, an accurate result can be achieved with only a few snapshots (previous methods require 20 or more).

\section{Geometry Constraints Formulation} \label{setupFeatures}
Our calibration setup is shown in Fig.~\ref{fig:calibTarget}. A V-shaped calibration target is formed by two triangular boards with a checkerboard on each triangle.
The angle between the two boards can be arbitrary as long as the two boards are not coplanar (the angle is 0 or 180 degree). In practice, the angle is set to arbitrarily obtuse to get good camera views of both two boards, and does not need to be known.
$P$, $Q$, $R$ and $O$ are four corners of the target.
We define the triangles as $T_1 = \bigtriangleup PQC$ and $T_2 = \bigtriangleup PRC$, and let $T_3 = \bigtriangleup PQO$  and $T_4 = \bigtriangleup PRO$.
For each observation, the scanning plane of the LRF intersects with the three sides $\overline{PQ}$, $\overline{PR}$ and $\overline{PO}$ at points ${^L}\mathbf{p}_1$, ${^L}\mathbf{p}_2$ and ${^L}\mathbf{p}_3$ respectively in the LRF frame. Moreover, the camera and LRF should be either synchronized or held stationary during data collection.
The camera is modeled by the standard pinhole model. We ignore lens distortions in the rest of the paper, and assume that the images have already been undistorted, e.g. using the functions from MATLAB Camera Calibration Toolbox~\cite{calibToolbox}.

Each observation of the calibration target consists of an image acquired from the camera and a single scan obtained from the LRF. The output of our calibration method is the relative transformation (${_L^C}\mathbf{R}$ and position ${^C}\mathbf{t}{_L}$) between the 2D LRF and the camera. As shown in Fig.~\ref{fig:calibTarget}, the input features from a single observation are: 1) three laser points ${^L}\mathbf{p}_1$, ${^L}\mathbf{p}_2$ and ${^L}\mathbf{p}_3$ from the LRF; 2) two unit normals ${^C}\mathbf{n}_1$ and ${^C}\mathbf{n}_2$ of perspective planes $T_1$, $T_2$ from the camera; 3) two unit normals ${^C}\mathbf{n}_3$ and ${^C}\mathbf{n}_4$ of board planes $T_3$ and $T_4$ in the camera frame, and two distances $d_1$ and $d_2$ from the camera to planes $T_3$ and $T_4$ respectively. Further details of feature extraction are described in Section~\ref{featureExtr}.

A single laser scan consists of a depth value for each angle at which the depth was sensed. In the LRF frame, we assume that the sensor is at its origin $L$. Let us express the feature points ${^L}\mathbf{p}_i$ as $[x_i, 0, z_i]^{\top}$, where $i = \{1, 2, 3\}$ are the indices of the feature points. Let the feature normals $\mathbf{n}_j$ of planes $T_j$ be $[a_j, b_j, c_j]^{\top}$, where $j = \{1, 2, 3, 4\}$. 
We now have a correspondence between a 3D point in LRF frame and a plane in camera frame. Thus, our constraint is that the laser point, transformed to the camera frame, must lie on the corresponding plane, which can be divided into three parts.

First, laser points ${^L}\mathbf{p}_1$ and ${^L}\mathbf{p}_2$ must lie on the planes $T_1$ and $T_2$, respectively. Then, the first two constraints have the form
\begin{equation} \label{constraint1}
{^C}\mathbf{n}_i^{\top} \left({_L^C}\mathbf{R} \cdot {^L}\mathbf{p}_i + {^C}\mathbf{t}{_L}\right) = 0 ,
\quad i = \{ 1, 2\}
\end{equation}
where ${_L^C}\mathbf{R} \in SO(3)$ and ${^C}\mathbf{t}{_L}$ are the unknowns. Second, for laser points ${^L}\mathbf{p}_1$ and ${^L}\mathbf{p}_3$, they must both lie on the plane $T_3$. Then, we have other two constraints
\begin{equation} \label{constraint2}
{^C}\mathbf{n}_3^{\top} \left({_L^C}\mathbf{R} \cdot {^L}\mathbf{p}_j + {^C}\mathbf{t}{_L}\right) = d_1 ,
\quad j = \{ 1, 3\} .
\end{equation}
Similarly, laser points ${^L}\mathbf{p}_2$ and ${^L}\mathbf{p}_3$ must both lie on the plane $T_4$. This gives two more constraints:
\begin{equation} \label{constraint3}
{^C}\mathbf{n}_4^{\top} \left({_L^C}\mathbf{R} \cdot {^L}\mathbf{p}_{k} + {^C}\mathbf{t}{_L}\right) = d_2 ,
\quad k = \{ 2, 3\}.
\end{equation}
As stated in Section~\ref{relatedWork}, once we solve for two columns $\mathbf{r}_1$ and $\mathbf{r}_3$ of ${_L^C}\mathbf{R}$, the second column $\mathbf{r}_2$ can be obtained by
\begin{equation} \label{crossProdR2}
\mathbf{r}_2 = \mathbf{r}_3 \times \mathbf{r}_1 .
\end{equation}
To summarize, we have nine unknowns (in $\mathbf{r}_1$, $\mathbf{r}_3$ and ${^C}\mathbf{t}{_L}$) and a system of six linear (Eqs.~(\ref{constraint1})-(\ref{constraint3})) and three nonlinear equations (Eq.~(\ref{constraint4})). In the next section, we show that these nine constraints are independent and hence sufficient to obtain a solution.

\section{Uniqueness Of The Solution} \label{uniqueSolution}
In this section, we prove that the features from a single observation of our calibration target constrain the calibration problem to a finite number of solutions.

For a single observation of the calibration target, our method builds up a system of Eqs.~(\ref{constraint4})-(\ref{constraint3}).
In order to prove the proposed method does not induce any ambiguity, the nine equations must be independent. We show that the first six linear equations are linearly independent. Since the other three nonlinear equations have no relationship with geometry constraints, they are independent from the first six linear equations.

From the constraints formulation, the six linear equations can be expressed as the following form
\begin{equation} \label{sixLinear}
\mathcal{A} \mathcal{X} = \mathcal{B} , \quad \mathcal{X} = [{^C}\mathbf{t}{_L}^{\top}, \mathbf{r}_1^{\top}, \mathbf{r}_3^{\top}]^{\top} ,
\end{equation}
where $\mathcal{X}$ is the vector of unknowns with ${^C}\mathbf{t}{_L} = [t_1, t_2, t_3]^{\top}$, $\mathbf{r}_1 = [r_{11}, r_{21}, r_{31}]^{\top}$ and $\mathbf{r}_3 = [r_{13}, r_{23}, r_{33}]^{\top}$, $\mathcal{B}$ is the distance vector denoted as $\mathcal{B} = [0, 0, d_1, d_1, d_2, d_2]^{\top}$, and $\mathcal{A}$ is the coefficient matrix whose elements are expressed using components from $^C\mathbf{n}_i$ and $^L\mathbf{p}_j$ as defined in Section~\ref{setupFeatures}. 
Lemma~\ref{lemma1} below states that the three unit vectors $^C\mathbf{n}_1$, $^C\mathbf{n}_2$ and $^C\mathbf{n}_3$ are linearly independent, which means they  span the entire 3D space.

\begin{lemma} \label{lemma1}
Suppose ${^C}\mathbf{n}_i$ is the normal vector of plane $T_i$ for $i = {1,2,3,4}$ as defined in Fig.~\ref{fig:calibTarget}, these normal vectors in any cardinality three subset of $\{{^C}\mathbf{n}_1, {^C}\mathbf{n}_2, {^C}\mathbf{n}_3, {^C}\mathbf{n}_4\}$ are linearly independent (totally four subsets: I. ${^C}\mathbf{n}_1$, ${^C}\mathbf{n}_2$ and ${^C}\mathbf{n}_3$; II. ${^C}\mathbf{n}_1$, ${^C}\mathbf{n}_2$ and ${^C}\mathbf{n}_4$; III. ${^C}\mathbf{n}_1$, ${^C}\mathbf{n}_3$ and ${^C}\mathbf{n}_4$; IV. ${^C}\mathbf{n}_2$, ${^C}\mathbf{n}_3$ and ${^C}\mathbf{n}_4$.).
\end{lemma} The proof is presented in Appendix~\ref{NonIndependence}.

As a corollary, the unit vector $^C\mathbf{n}_4$ can be expressed as the combination of first three unit vectors $^C\mathbf{n}_4 = u\cdot ^C\mathbf{n}_1 + v\cdot ^C\mathbf{n}_2 + w\cdot ^C\mathbf{n}_3$.
This allows us to perform Gaussian elimination on Eq.~(\ref{sixLinear}) as follows:
\begin{itemize}
\item Keep $\textit{\textbf{row}}_1$ and $\textit{\textbf{row}}_2$ unchanged, and let $\textit{\textbf{Row}}_3 \leftarrow \textit{\textbf{row}}_4$;
\item Let $\textit{\textbf{Row}}_4 \leftarrow \textit{\textbf{row}}_3 - \textit{\textbf{row}}_4$ and $\textit{\textbf{Row}}_5 \leftarrow \textit{\textbf{row}}_6 - \textit{\textbf{row}}_5$;
\item Let $\textit{\textbf{Row}}_6 \leftarrow \textit{\textbf{row}}_5 - (u\cdot \textit{\textbf{row}}_1 + v\cdot \textit{\textbf{row}}_2 + w\cdot \textit{\textbf{row}}_3)$.
\end{itemize}
Here, $\mathcal{A}$ is transformed as
\begin{equation}
\mathcal{A} = \scalebox{0.55}
{$
\begin{bmatrix}
a_1 & b_1 & c_1 & a_1x_1 & b_1x_1 & c_1x_1 & a_1z_1 & b_1z_1 & c_1z_1 \\
a_2 & b_2 & c_2 & a_2x_2 & b_2x_2 & c_2x_2 & a_2z_2 & b_2z_2 & c_2z_2 \\
a_3 & b_3 & c_3 & a_3x_1 & b_3x_1 & c_3x_1 & a_3z_1 & b_3z_1 & c_3z_1 \\
a_3 & b_3 & c_3 & a_3x_3 & b_3x_3 & c_3x_3 & a_3z_3 & b_3z_3 & c_3z_3 \\
a_4 & b_4 & c_4 & a_4x_2 & b_4x_2 & c_4x_2 & a_4z_2 & b_4z_2 & c_4z_2 \\
a_4 & b_4 & c_4 & a_4x_3 & b_4x_3 & c_4x_3 & a_4z_3 & b_4z_3 & c_4z_3
\end{bmatrix}
$}
\rightarrow
\bar{\mathcal{A}} = \scalebox{0.6}
{$
\begin{bmatrix}
\mathcal{P}_{\alpha} & \mathcal{Q}_{\alpha} \\
\mathbf{0}_{3 \times 3} & \bar{\mathcal{P}}_{\beta} | \bar{\mathcal{Q}}_{\beta}
\end{bmatrix}
$}
\end{equation}
with sub-matrices $\mathcal{P}_{\alpha}$, $\bar{\mathcal{P}}_{\beta}$, $\mathcal{Q}_{\alpha}$ and $\bar{\mathcal{Q}}_{\beta}$ where
\begin{equation}
\begin{split}
\mathcal{P}_{\alpha} &= \scalebox{0.7}
{$
\begin{bmatrix}
a_1 & b_1 & c_1 \\
a_2 & b_2 & c_2 \\
a_3 & b_3 & c_3
\end{bmatrix}
$} \\
\bar{\mathcal{P}}_{\beta} &= \scalebox{0.7}
{$
\begin{bmatrix}
a_3 & b_3 & c_3 \\
a_4 & b_4 & c_4 \\
a_{13} & b_{13} & c_{13}
\end{bmatrix}
$}
\end{split} ,
\begin{split}
\mathcal{Q}_{\alpha} &= \scalebox{0.7}
{$
\begin{bmatrix}
a_1x_1 & b_1x_1 & c_1x_1 & a_1z_1 & b_1z_1 & c_1z_1 \\
a_2x_2 & b_2x_2 & c_2x_2 & a_2z_2 & b_2z_2 & c_2z_2 \\
a_3x_3 & b_3x_3 & c_3x_3 & a_3z_3 & b_3z_3 & c_3z_3
\end{bmatrix}
$} \\
\bar{\mathcal{Q}}_{\beta} &= \scalebox{0.7}
{$
\begin{bmatrix}
a_3k_{\gamma} & b_3k_{\gamma} & c_3k_{\gamma} \\
a_4k_{\beta} & b_4k_{\beta} & c_4k_{\beta} \\
a_{13}k_{\alpha} & b_{13}k_{\alpha} & c_{13}k_{\alpha}
\end{bmatrix}
$}
\end{split}
\end{equation}
with $k_{\alpha} = \dfrac{z_1-z_2}{x_1-x_2}$, $k_{\beta} = \dfrac{z_2-z_3}{x_2-x_3}$, $k_{\gamma} = \dfrac{z_3-z_1}{x_3-x_1}$ and $a_{13} = ua_1 + wa_3$, $b_{13} = ub_1 + wb_3$, $c_{13} = uc_1 + wc_3$. We perform one more step of Gaussian elimination:
\begin{itemize}
\item Let $\textit{\textbf{Row}}_6 \leftarrow \textit{\textbf{Row}}_6 - w\cdot \textit{\textbf{Row}}_4$.
\end{itemize}
The matrix $\bar{\mathcal{A}}$ is transformed as
\begin{equation}
\bar{\mathcal{A}} ~
\rightarrow ~
\tilde{\mathcal{A}} = \scalebox{0.6}
{$
\begin{bmatrix}
\mathcal{P}_{\alpha} & \mathcal{Q}_{\alpha} \\
\mathbf{0}_{3 \times 3} & \mathcal{P}_{\beta}|\mathcal{Q}_{\beta}
\end{bmatrix}
$}
\end{equation}
with sub-matrices $\mathcal{P}_{\beta}$ and $\mathcal{Q}_{\beta}$ where
\begin{equation}
\begin{split}
\mathcal{P}_{\beta} &= \scalebox{0.7}
{$
\begin{bmatrix}
a_3 & b_3 & c_3 \\
a_4 & b_4 & c_4 \\
a_1 & b_1 & c_1
\end{bmatrix}
$}
\end{split} ,
\begin{split}
\mathcal{Q}_{\beta} &= \scalebox{0.7}
{$
\begin{bmatrix}
a_3k_{\gamma} & b_3k_{\gamma} & c_3k_{\gamma} \\
a_4k_{\beta} & b_4k_{\beta} & c_4k_{\beta} \\
a_1k_{\alpha}+a_3k_{\delta} & b_1k_{\alpha}+b_3k_{\delta} & c_1k_{\alpha}+c_3k_{\delta}
\end{bmatrix}
$}
\end{split}
\end{equation}
with $k_{\delta} = \left[ \dfrac{w}{u} \cdot a_3(k_{\alpha}-k_{\gamma}) \right]$. Since laser features ${^L}\mathbf{p}_1$, ${^L}\mathbf{p}_2$ and ${^L}\mathbf{p}_3$ are extracted from two distinct line segments, 
their $X_L$ coordinates cannot be equal otherwise these three points are on a same plane from an invalid observation. Therefore, $k_{\alpha}$, $k_{\beta}$ and $k_{\gamma}$ can be calculated. After Gaussian elimination, the distance vector $\mathcal{B}$ is transformed into a new vector denoted as $\tilde{\mathcal{B}} = [0, 0, d_1, 0, 0, \tilde{d}]^{\top}$, where $\tilde{d} = \frac{d_2-wd_1}{u(x_2-x_1)}$.

Let us first take a close look at the structure of $\tilde{\mathcal{A}}$. Since we know that unit vectors $^C\mathbf{n}_1$, $^C\mathbf{n}_2$ and $^C\mathbf{n}_3$ are linearly independent (Lemma~\ref{lemma1}), matrix $\mathcal{P}_{\alpha}$ is non-singular such that we can reduce it to an upper triangular matrix. Thus, the first three linear equations are independent. Next, the unit vectors $^C\mathbf{n}_1$, $^C\mathbf{n}_3$ and $^C\mathbf{n}_4$ are also linearly independent (Lemma~\ref{lemma1}). Then, matrix $\mathcal{P}_{\beta}$ is also non-singular and can be reduced into an upper triangular matrix, which means the last three linear equations are also independent. From the procedure above, we have just reduced the $\tilde{\mathcal{A}}$ to a matrix which has a lower triangular corner with zero elements, just shown as follows
\begin{equation} \label{matrixStructure}
\tilde{\mathcal{A}} \rightarrow
\begin{bmatrix}
\bigtriangledown_{3 \times 3} &\Box_{3 \times 3} &\Box_{3 \times 3} \\
\mathbf{0}_{3 \times 3} &\bigtriangledown_{3 \times 3} &\Box_{3 \times 3}
\end{bmatrix} ,
\end{equation}
where $\bigtriangledown$ represents a $3 \times 3$ upper triangular matrix and $\Box$ represents a $3 \times 3$ square matrix. From the matrix structure in Eq.~(\ref{matrixStructure}), we can conclude that the six linear equations for geometry constraints are linearly independent, which means plus the other three nonlinear equations we can solve for nine unknown components in $\mathbf{r}_1$, $\mathbf{r}_3$ and ${^C}\mathbf{t}{_L}$, respectively.
Hence, there is no ambiguity in our proposed method in which the relative pose between the LRF and the camera is determined from a single snapshot of the calibration target. \QEDA

\section{Analytical Solution} \label{solutionTech}
In this section, we first present how to obtain the solution for the extrinsic calibration of the LRF-camera system from just a single observation of the calibration target. Then, we show the solution from multiple observations which is needed to reduce the effect of noise. Note that an analytical solution is obtained in our constraints system, which is more general than the closed-form solution in~\cite{Naroditsky}. Moreover, 
we present a strategy to exclude invalid solutions from the cheirality check.

\subsection{From a Single Observation} \label{closedForm}
We outline seven steps to solve the polynomial system (Eqs.~(\ref{constraint4})-(\ref{constraint3})).
For convenience, the geometry constraints (Eqs.~(\ref{constraint1})-(\ref{constraint3})) are reformulated as follows
\begin{equation} \label{constraintRe}
{^C}\bar{\mathbf{n}}_i^{\top} \left({_L^C}\mathbf{R} \cdot {^L}\bar{\mathbf{p}}_i + {^C}\mathbf{t}{_L} \right) = \bar{d}_i, \quad i = 1,2,...,6
\end{equation}
where the parameters ${^C}\bar{\mathbf{n}}_i$, ${^L}\bar{\mathbf{p}}_i$ and $\bar{d}_i$ are defined as
\begin{equation} \label{reDefinition}
\scalebox{0.7}
{$
\begin{cases}
{^C}\bar{\mathbf{n}}_i = {^C}\mathbf{n}_i, &i = 1,2 \\
{^C}\bar{\mathbf{n}}_j = {^C}\mathbf{n}_3, &j = 3,4 \\
{^C}\bar{\mathbf{n}}_k = {^C}\mathbf{n}_4, &k = 5,6
\end{cases}
$} , \ 
\scalebox{0.7}
{$
\begin{cases}
{^L}\bar{\mathbf{p}}_i = {^L}\mathbf{p}_1, &i = 1,3 \\
{^L}\bar{\mathbf{p}}_j = {^L}\mathbf{p}_2, &j = 2,5 \\
{^L}\bar{\mathbf{p}}_k = {^L}\mathbf{p}_3, &k = 4,6
\end{cases}
$} , \ 
\scalebox{0.7}
{$
\begin{cases}
\bar{d}_i = 0, &i = 1,2 \\
\bar{d}_j = d_1, &j = 3,4 \\
\bar{d}_k = d_2, &k = 5,6
\end{cases}
$} .
\end{equation}

\noindent
\textbf{STEP 1:} The problem is reformulated in the view of nonlinear optimization as stated in~\cite{Guo} shown below
\begin{equation} \label{nonlinearRe}
\begin{gathered}
\argmin_{{_L^C}\mathbf{R}, {^C}\mathbf{t}{_L}} J = \sum_{i=1}^N \left( {^C}\bar{\mathbf{n}}_i^{\top} \left({_L^C}\mathbf{R} \cdot {^L}\bar{\mathbf{p}}_i + {^C}\mathbf{t}{_L} \right) - \bar{d}_i \right)^2 \\
\text{s. t.} \quad {_L^C}\mathbf{R}^{\top} \cdot {_L^C}\mathbf{R} = \mathbf{I}, \quad \det \left({_L^C}\mathbf{R}\right) = 1
\end{gathered} ,
\end{equation}
where $N = 6$.
From the reformulated problem~(\ref{nonlinearRe}),
the optimal solution for ${^C}\mathbf{t}{_L}$ is obtained
as shown below
\begin{equation} \label{translation}
\begin{gathered}
\dfrac{\partial J}{\partial{^C}\mathbf{t}{_L}} = \sum_{i=1}^N 2\left[ {^C}\bar{\mathbf{n}}_i^{\top} \left( {_L^C}\mathbf{R} \cdot {^L}\bar{\mathbf{p}}_i + {^C}\mathbf{t}{_L} \right) - \bar{d}_i \right] {^C}\bar{\mathbf{n}}_i = 0 \\
\Rightarrow {^C}\mathbf{t}{_L} = \mathcal{N}_o^{-1} \left( \mathcal{D}_n - \sum_{i=1}^N {^C}\bar{\mathbf{n}}_i {^C}\bar{\mathbf{n}}_i^{\top} {_L^C}\mathbf{R} {^L}\bar{\mathbf{p}}_i \right) ,
\end{gathered}
\end{equation}
where $\mathcal{N}_o = \sum_{i=1}^N {^C}\bar{\mathbf{n}}_i {^C}\bar{\mathbf{n}}_i^{\top}$ and $\mathcal{D}_n = \sum_{i=1}^N \bar{d}_i {^C}\bar{\mathbf{n}}_i$.
\begin{lemma} \label{lemma2}
$\mathcal{N}_o$ is a non-singular matrix and thus invertible.
\end{lemma}

\noindent
Lemma~\ref{lemma2} is proved in Appendix~\ref{NonSingular}.

Since a laser point is defined as ${^L}\bar{\mathbf{p}}_i = [\bar{x}_i, 0, \bar{z}_i]^{\top}$, we arrange the expression of ${^C}\mathbf{t}{_L}$ in~(\ref{translation}) to the form
\begin{equation} \label{translationS}
{^C}\mathbf{t}{_L} = \mathcal{N}_o^{-1} \left(\mathcal{D}_n - \mathcal{N}_{\alpha}\mathbf{r}_1 -  \mathcal{N}_{\gamma}\mathbf{r}_3 \right) ,
\end{equation}
in which $\mathcal{N}_{\alpha} = \sum_{i=1}^N {^C}\bar{\mathbf{n}}_i {^C}\bar{\mathbf{n}}_i^{\top} \bar{x}_i$ and $\mathcal{N}_{\gamma} = \sum_{i=1}^N {^C}\bar{\mathbf{n}}_i {^C}\bar{\mathbf{n}}_i^{\top} \bar{z}_i$.

\noindent
\textbf{STEP 2:} With $\mathcal{N}_x$, $\mathcal{N}_z$, $\mathcal{N}$ and $\mathcal{D}$ defined as
\begin{equation}
\mathcal{N}_x = \scalebox{0.7}
{$
\begin{bmatrix}
\bar{x}_1{^C}\bar{\mathbf{n}}_1^{\top}\\
\vdots \\
\bar{x}_N{^C}\bar{\mathbf{n}}_N^{\top}
\end{bmatrix}
$} , \ 
\mathcal{N}_z = \scalebox{0.7}
{$
\begin{bmatrix}
\bar{z}_1{^C}\bar{\mathbf{n}}_1^{\top} \\
\vdots \\
\bar{z}_N{^C}\bar{\mathbf{n}}_N^{\top}
\end{bmatrix}
$} , \ 
\mathcal{N} = \scalebox{0.7}
{$
\begin{bmatrix}
{^C}\bar{\mathbf{n}}_1^{\top} \\
\vdots \\
{^C}\bar{\mathbf{n}}_N^{\top}
\end{bmatrix}
$} , \ 
\mathcal{D} = \scalebox{0.7}
{$
\begin{bmatrix}
\bar{d}_1\\
\vdots \\
\bar{d}_N
\end{bmatrix}
$} ,
\end{equation}
we substitute~(\ref{translationS}) in constraints~(\ref{constraintRe}) and obtain
\begin{equation} \label{expressionR1P}
\mathcal{G}_x \mathbf{r}_1 + \mathcal{G}_z \mathbf{r}_3 = \mathcal{G}_d ,
\end{equation}
where $\mathcal{G}_x = \mathcal{N}_x-\mathcal{N}\mathcal{N}_o^{-1}\mathcal{N}_{\alpha}$, $\mathcal{G}_z = \mathcal{N}_z-\mathcal{N}\mathcal{N}_o^{-1}\mathcal{N}_{\gamma}$ and $\mathcal{G}_d = \mathcal{D}-\mathcal{N}\mathcal{N}_o^{-1}\mathcal{D}_n$.
Then, $\mathbf{r}_1$ is further expressed in terms of $\mathbf{r}_3$
\begin{equation} \label{expressionR1}
\mathbf{r}_1 = \mathcal{H}\mathbf{r}_3 + \mathcal{K} ,
\end{equation}
where $\mathcal{H} = - \left(\mathcal{G}_x^{\top}\mathcal{G}_x\right)^{-1}\mathcal{G}_x^{\top}\mathcal{G}_z$ and $\mathcal{K} = \left(\mathcal{G}_x^{\top}\mathcal{G}_x\right)^{-1}\mathcal{G}_x^{\top}\mathcal{G}_d$.

Note that $\mathcal{G}_x^{\top}\mathcal{G}_x$ is invertible. The proof is by contradiction. We first assume the $3 \times 3$ matrix $\mathcal{G}_x^{\top}\mathcal{G}_x$ is non-invertible thus rank deficient. From~(\ref{expressionR1P}), we have $\left(\mathcal{G}_x^{\top}\mathcal{G}_x\right) \mathbf{r}_1 = \mathcal{G}_x^{\top} \mathcal{G}_d - \mathcal{G}_x^{\top}\mathcal{G}_z \mathbf{r}_3$, which is a $\mathbf{A}\mathbf{x} = \mathbf{b}$ system for solving $\mathbf{r}_1$. Then for any given $\mathbf{r}_3$ (thus $\mathbf{b}$ is given), the rank-deficient $\mathbf{A}$ results in infinite solutions for $\mathbf{x}$ (which is $\mathbf{r}_1$). It means that our system has ambiguity, which is in contradiction to the uniqueness proof in Section~\ref{uniqueSolution}. Hence, $\mathcal{G}_x^{\top}\mathcal{G}_x$ is invertible. \QEDA

\noindent
\textbf{STEP 3:} Now we can eliminate $\mathbf{r}_1$ by substituting~(\ref{expressionR1}) in the three remaining second order constraints~(\ref{constraint4}). After full expansion, we have the following
\begin{align}
\scalebox{0.55}
{$
e_{11}r_{13}^2 + e_{12}r_{13}r_{23} + e_{13}r_{23}^2 + e_{14}r_{13}r_{33} + e_{15}r_{23}r_{33} + e_{16}r_{33}^2 + e_{17}r_{13} + e_{18}r_{23} + e_{19}r_{33} + m = 0
$} \label{constraint1R} \\
\scalebox{0.55}
{$
e_{21}r_{13}^2 + e_{22}r_{13}r_{23} + e_{23}r_{23}^2 + e_{24}r_{13}r_{33} + e_{25}r_{23}r_{33} + e_{26}r_{33}^2 + e_{27}r_{13} + e_{28}r_{23} + e_{29}r_{33} = 0
$} \label{constraint2R} \\
r_{13}^2+r_{23}^2+r_{33}^2 - 1 = 0 , \label{objectR}
\end{align}
where the coefficients $e_{ij}$ and the constant $m$ are computed in a closed form in terms of the components of $\mathcal{H}$ and $\mathcal{K}$.
To solve the polynomial system (Eqs.~(\ref{constraint1R})-(\ref{objectR})), we aim to obtain a univariate polynomial in $r_{33}$ using Macaulay resultant~\cite{Macaulay}. This multivariate resultant is the ratio of two determinants, the denominator~(\ref{resDenominator}) and numerator~(\ref{resNumerator})
\begin{equation} \label{resDenominator}
\begin{bmatrix}
e_{11} & 0 & 0 \\
0 & e_{11} & 1 \\
E_{16} & e_{13} & 1
\end{bmatrix} , \quad E_{16} = e_{16}r_{33}^2 + e_{19}r_{33} + m
\end{equation}
\begin{equation} \label{resNumerator}
\scalebox{0.55}
{$
\begin{bmatrix}
e_{11} &0      &0      &0      &0      &0      &0      &0      &0      &0      &0      &0      &0      &0      &0      \\
e_{12} &e_{11} &0      &0      &0      &0      &1      &0      &e_{21} &0      &0      &0      &0      &0      &0      \\
E_{14} &0      &e_{11} &0      &0      &0      &0      &1      &0      &e_{21} &0      &0      &0      &0      &0      \\
e_{13} &e_{12} &0      &e_{11} &0      &0      &0      &0      &e_{22} &0      &1      &0      &0      &0      &0      \\
E_{15} &E_{14} &e_{12} &0      &e_{11} &0      &0      &0      &E_{24} &e_{22} &0      &1      &0      &e_{21} &0      \\
E_{16} &0      &E_{14} &0      &0      &e_{11} &0      &0      &0      &E_{24} &0      &0      &1      &0      &e_{21} \\
0 	   &e_{13} &0      &e_{12} &0      &0      &1      &0      &e_{23} &0      &0      &0      &0      &0      &0      \\
0 	   &E_{15} &e_{13} &E_{14} &e_{12} &0      &0      &1      &E_{25} &e_{23} &0      &0      &0      &e_{22} &0      \\
0 	   &E_{16} &E_{15} &0      &E_{14} &e_{12} &E_{31} &0      &E_{26} &E_{25} &0      &0      &0      &E_{24} &e_{22} \\
0 	   &0      &E_{16} &0      &0      &E_{14} &0      &E_{31} &0      &E_{26} &0      &0      &0      &0      &E_{24} \\
0      &0      &0      &e_{13} &0      &0      &0      &0      &0      &0      &1      &0      &0      &0      &0      \\
0      &0      &0      &E_{15} &e_{13} &0      &0      &0      &0      &0      &0      &1      &0      &e_{23} &0      \\
0      &0      &0      &E_{16} &E_{15} &e_{13} &0      &0      &0      &0      &E_{31} &0      &1      &E_{25} &e_{23} \\
0      &0      &0      &0      &E_{16} &E_{15} &0      &0      &0      &0      &0      &E_{31} &0      &E_{26} &E_{25} \\
0      &0      &0      &0      &0      &E_{16} &0      &0      &0      &0      &0      &0      &E_{31} &0      &E_{26} 
\end{bmatrix}
$} ,
\end{equation}
with elements $E_{14}$, $E_{15}$, $E_{31}$, $E_{24}$, $E_{25}$ and $E_{26}$ defined as
\begin{equation}
\scalebox{0.85}
{$
\begin{aligned}
E_{14} &= e_{14}r_{33} + e_{17}, \ E_{15} = e_{15}r_{33} + e_{18}, \ E_{31} = r_{33}^2 - 1 \\
E_{24} &= e_{24}r_{33} + e_{27}, \ E_{25} = e_{25}r_{33} + e_{28}, \ E_{26} = e_{26}r_{33}^2 + e_{29}r_{33}
\end{aligned}
$} .
\end{equation}
We set this resultant to $0$, and obtain the univariate polynomial equation
\begin{equation}
\scalebox{0.75}
{$
g_1r_{33}^8 + g_2r_{33}^7 + g_3r_{33}^6 + g_4r_{33}^5 + g_5r_{33}^4 + g_6r_{33}^3 + g_7r_{33}^2 + g_8r_{33} + g_9 = 0
$} ,
\end{equation}
where the coefficients $g_i$ are computed in closed-form of the coefficients of Eqs.~(\ref{constraint1R})-(\ref{objectR}). 

Although the eighth-degree (higher than four) univariate polynomial $\mathbf{P}$ does not have a closed-form roots expression, we can obtain its all roots by computing the eigenvalues of its companion matrix $\mathcal{C}(\mathbf{P})$~\cite{companionM}. For numerical stability, we approximate the roots through an iterated method~\cite{companionIt} which uses a generalized companion matrix $\mathcal{C}(\mathbf{P}, \mathbf{S})$ constructed from $\mathbf{P}$ and initialized by $\mathcal{C}(\mathbf{P})$. Here, $\mathcal{C}(\mathbf{P})$ and $\mathcal{C}(\mathbf{P}, \mathbf{S})$ are expressed as
\begin{equation}
\scalebox{0.55}
{$
\mathcal{C}(\mathbf{P}) =
\begin{bmatrix}
0 & 1 & 0 & \cdots & 0 \\
0 & 0 & 1 & \cdots & 0 \\
\vdots & \vdots & \vdots & \ddots & \vdots \\
-\frac{g_9}{g_1} & -\frac{g_8}{g_1} & -\frac{g_7}{g_1} & \cdots & -\frac{g_2}{g_1}
\end{bmatrix}
$} ,
\scalebox{0.55}
{$
\mathcal{C}(\mathbf{P}, \mathbf{S}) =
\begin{bmatrix}
s_1 & 0 & \cdots & 0 \\
0 & s_2 & \cdots & 0 \\
\vdots & \vdots & \ddots & \vdots \\
0 & 0 & \cdots & s_8
\end{bmatrix} - 
\begin{bmatrix}
l_1 & l_2 & \cdots & l_8 \\
l_1 & l_2 & \cdots & l_8 \\
\vdots & \vdots & \vdots & \vdots \\
l_1 & l_2 & \cdots & l_8
\end{bmatrix}
$} ,
\end{equation}
where $\mathbf{S} = (s_1, ... , s_8)$, $l_i = \lvert \frac{\mathbf{P}(s_i)}{\mathbf{Q}^{\prime}(s_i)} \rvert$ and $\mathbf{Q}^{\prime}(s_i) = \prod_{i \neq j}(s_i - s_j)$. $\mathbf{S}$ is first initialized as the eigenvalues of $\mathcal{C}(\mathbf{P})$. Then for each iteration, $\mathbf{S}$ is updated as the eigenvalues of $\mathcal{C}(\mathbf{P}, \mathbf{S})$ until $\mathbf{S}$ is converged. Eight possible solutions for $r_{33}$ are obtained.

\noindent
\textbf{STEP 4:} Each solution for $r_{33}$ (numeric value $\hat{r_{33}}$) corresponds to a single solution for the rest of the unknowns. For numerical stability, we compute the Sylvester resultant~\cite{Macaulay} of Eq.~(\ref{constraint1R}) and Eq.~(\ref{objectR}) w.r.t. $r_{23}$. With the determinant of this resultant set to zero, we obtain a quartic polynomial $\mathbf{P}_1$ in $r_{13}$
\begin{equation}
\scalebox{0.6}
{$
\mathbf{P}_1 = \det \left( \begin{bmatrix}
f_{12} & e_{12}r_{13} + e_{15}\hat{r_{33}} + e_{18} & e_{13} & 0 \\
0 & f_{12} & e_{12}r_{13} + e_{15}\hat{r_{33}} + e_{18} & e_{13} \\
r_{13}^2 + \hat{r_{33}}^2 - 1 & 0 & 1 & 0 \\
0 & r_{13}^2 + \hat{r_{33}}^2 - 1 & 0 & 1
\end{bmatrix} \right)  = 0
$} ,
\end{equation}
where $f_{12} = e_{11}r_{13}^2 + e_{14}r_{13}\hat{r_{33}} + e_{16}\hat{r_{33}}^2 + e_{17}r_{13} + e_{19}\hat{r_{33}} + m$. Similarly, we compute the Sylvester resultant of Eq.~(\ref{constraint2R}) and Eq.~(\ref{objectR}) w.r.t. $r_{23}$, and set its determinant to zero to obtain another quartic polynomial $\mathbf{P}_2$ in $r_{13}$.  To solve this overdetermined system, we aim to minimize the sum of the squares of $\mathbf{P}_1$ and $\mathbf{P}_2$, and thus set the derivative of $\mathbf{P}_1^2$ + $\mathbf{P}_2^2$ w.r.t. $r_{13}$ to zero to get a seventh-degree polynomial. Seven solutions for $r_{13}$ obtained by iterated method mentioned above are tested if both $\mathbf{P}_1(\hat{r_{13}}) = 0$ and $\mathbf{P}_2(\hat{r_{13}}) = 0$.

After substituting the numeric solutions $\hat{r_{13}}$ and $\hat{r_{33}} $ into Eqs.~(\ref{constraint1R})-(\ref{objectR}), we perform the same optimization method to solve the overdetermined system for $r_{23}$. Note that we have a closed-form roots expression for the third-degree polynomial obtained from the derivative of the cost function w.r.t. $r_{23}$ (the sum of the squares of three polynomials in (\ref{constraint1R})-(\ref{objectR})). We only keep the solution $\hat{r_{23}}$ if all Eqs.~(\ref{constraint1R})-(\ref{objectR}) hold.

\noindent
\textbf{STEP 5:} After obtaining $\mathbf{r}_{3}$, $\mathbf{r}_{1}$ can be calculated from Eq.~(\ref{expressionR1}) and $\mathbf{r}_{2}$ can be retrieved from Eq.~(\ref{crossProdR2}). Finally, ${^C}\mathbf{t}{_L}$ can be obtained using Eq.~(\ref{translation}).

Eight possible solutions give us up to four real solutions. Four complex solutions can be eliminated as follows. We square all the elements of $\mathbf{r}_{1}$, $\mathbf{r}_{2}$, $\mathbf{r}_{3}$ and ${^C}\mathbf{t}{_L}$, and check if they all have non-negative real parts.

\noindent
\textbf{STEP 6:} In practice, while the solution for $\mathbf{r}_3$ fails to deliver a real solution in the presence of noise, we use its projection on real domain as the initial value. Eqs.~(\ref{constraint1R})-(\ref{objectR}) are then treated as a whole $\mathcal{F}(\mathbf{x}) = \mathbf{0}$ for $\mathbf{r}_3$, which can be solved using the Trust-Region Dogleg method~\cite{Conn}~\cite{Powell}. At each iteration $k$, the trust region subproblem here is
\begin{equation} \label{trustRegionDogleg}
\begin{gathered}
\argmin_{\mathbf{d}_k} \scalebox{0.7}
{$
\dfrac{1}{2}\mathcal{F}(\mathbf{x}_k)^{\top}\mathcal{F}(\mathbf{x}_k) + \mathbf{d}_k^{\top}\mathcal{J}(\mathbf{x}_k)^{\top}\mathcal{F}(\mathbf{x}_k) + \dfrac{1}{2}\mathbf{d}_k\mathcal{J}(\mathbf{x}_k)^{\top}\mathcal{J}(\mathbf{x}_k)\mathbf{d}_k
$} \\
\text{s. t. } \quad \Vert \mathbf{d}_k \Vert \le \Delta_k
\end{gathered} ,
\end{equation}
where $\Delta_k$ is updated, and the Jacobian is defined as $\mathcal{J}(\mathbf{x}_k) = \left[ \nabla\mathcal{F}_1(\mathbf{x}_k), \nabla\mathcal{F}_2(\mathbf{x}_k), \nabla\mathcal{F}_3(\mathbf{x}_k) \right]^{\top}$. The step $\mathbf{d}_k$ to obtain $\mathbf{x}_{k+1}$ is then constructed as 
\begin{equation}
\mathbf{d}_k = 
\begin{cases}
\lambda \mathbf{d}_C, &0 \le \lambda \le 1 \\
\mathbf{d}_C + (\lambda - 1)(\mathbf{d}_{GN} - \mathbf{d}_C),  &1 \le \lambda \le 2
\end{cases} ,
\end{equation}
where $\lambda$ is the largest value such that $\Vert \mathbf{d}_k \Vert \le \Delta_k$. With $\mathbf{g}_k = \mathcal{J}(\mathbf{x}_k)^{\top}\mathcal{F}(\mathbf{x}_k)$ and $\mathbf{B}_k = \mathcal{J}(\mathbf{x}_k)^{\top}\mathcal{J}(\mathbf{x}_k)$, the Cauchy step and the Gauss-Newton step are respectively defined as
$\mathbf{d}_C = - \frac{\mathbf{g}_k^{\top} \mathbf{g}_k}{\mathbf{g}_k^{\top} \mathbf{B}_k \mathbf{g}_k} \mathbf{g}_k$ and $\mathbf{d}_{GN} = - \mathbf{B}_k^{-1} \mathbf{g}_k$.

\begin{figure}[tb]
	\centering
	\includegraphics[width=0.9\columnwidth]{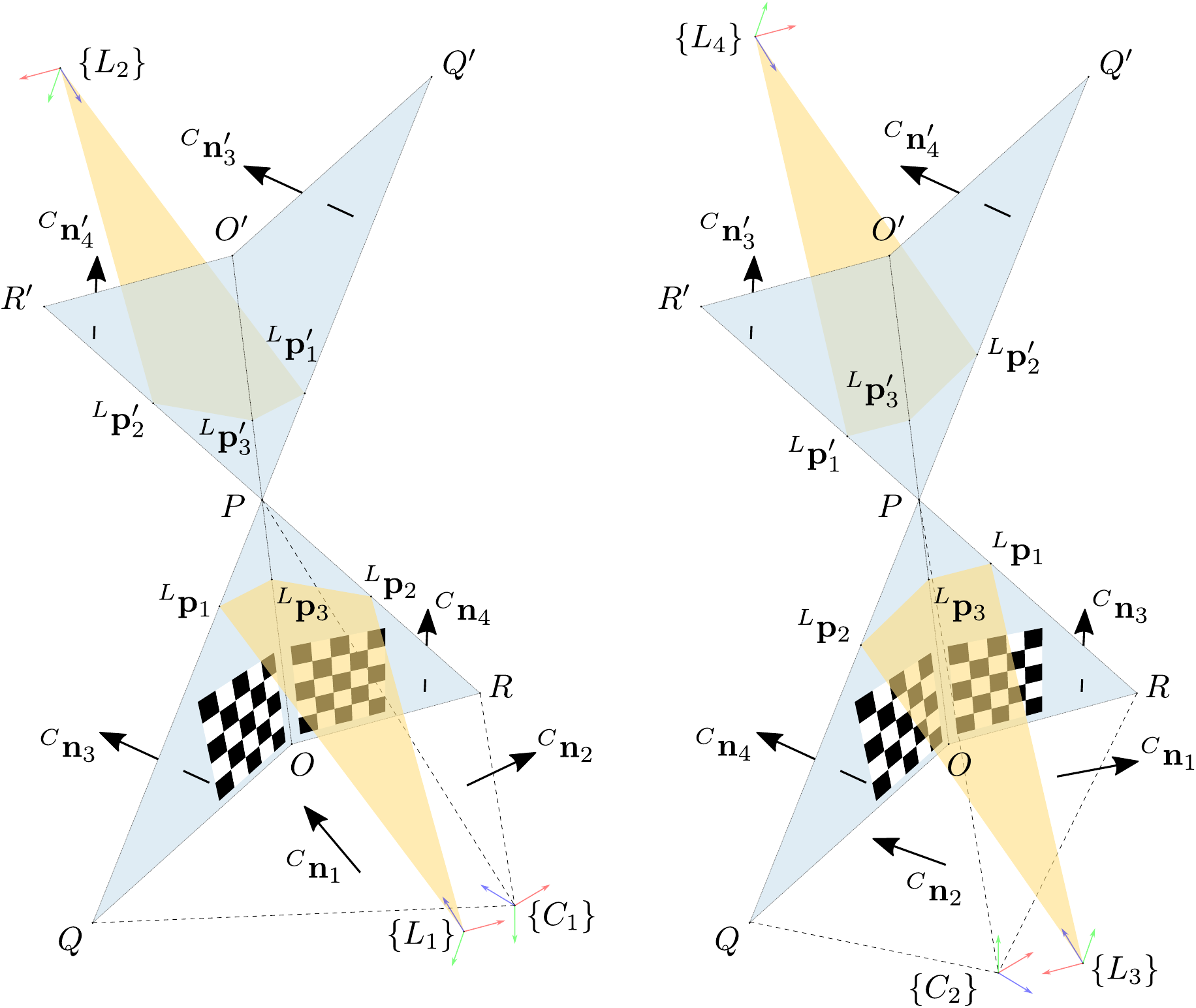}
	\caption{Four possible real solutions are shown as a combination of the LRF frame and the camera frame ($\{L_1, C_1\}$, $\{L_2, C_1\}$, $\{L_3, C_2\}$ and $\{L_4, C_2\}$). The unique solution $\{L_1, C_1\}$ is determined through the cheirality check.}
	\label{fig:cheiralityCheck}
	\vspace*{-1mm}
\end{figure}

\noindent
\textbf{STEP 7:} We must choose the unique solution from invalid solutions through the cheirality check. Fig.~\ref{fig:cheiralityCheck} shows four possible real solutions, which are $\{L_1, C_1\}$, $\{L_2, C_1\}$, $\{L_3, C_2\}$ and $\{L_4, C_2\}$ in terms of a combination of the LRF frame and the camera frame.

The unique solution $\{L_1, C_1\}$ is determined as follows. Both the LRF and the camera must face towards the same direction, and three laser points ${^L}\mathbf{p}_1$, ${^L}\mathbf{p}_2$ and ${^L}\mathbf{p}_3$ after transformation to the camera frame must be in front of the camera. These two conditions are
\begin{equation}
\begin{gathered}
\mathbf{n}_z^{\top} \cdot {_{L_1}^{C_1}}\mathbf{R} \cdot \mathbf{n}_z > 0, \quad \mathbf{n}_z = [0,0,1]^{\top} \\
\mathbf{n}_z^{\top} \cdot \left({_{L_1}^{C_1}}\mathbf{R} \cdot {^{L_1}}\mathbf{p}_i + {^{C_1}}\mathbf{t}{_{L_1}}\right) > 0, \quad i = \{ 1, 2, 3\}
\end{gathered} .
\end{equation}

\subsection{From Multiple Observations} \label{closedFormM}
Multiple observations from different views can be used to 
suppress noise.
The problem in this case is formulated as follows
\begin{equation} \label{multMonlinearObj}
\begin{gathered}
\argmin_{{_L^C}\mathbf{R}, {^C}\mathbf{t}{_L}} \sum_{i=1}^M \sum_{j=1}^N \left( {^C}\bar{\mathbf{n}}_{ij}^{\top} \left({_L^C}\mathbf{R} \cdot {^L}\bar{\mathbf{p}}_{ij} + {^C}\mathbf{t}{_L} \right) - \bar{d}_{ij} \right)^2 \\
\text{s. t. } \quad {_L^C}\mathbf{R}^{\top} \cdot {_L^C}\mathbf{R} = \mathbf{I}, \quad \det \left({_L^C}\mathbf{R}\right) = 1
\end{gathered} ,
\end{equation}
where $N = 6$ and $M$ is the total number of different multiple observations. The solution is obtained through the seven steps mentioned in Subsection~\ref{closedForm}. As the rotation is represented by the Rodrigues' formula, the calibration result is further refined using Levenberg-Marquardt (LM) method~\cite{Levenberg}~\cite{Marquardt}. Our solution in~(\ref{multMonlinearObj}) serves as the accurate initial value.

\section{Details of the Calibration Approach} \label{featureExtr}
We explain below how to accurately extract the features required for our method. These are four normal vectors ${^C}\mathbf{n}_i$ ($i = 1,2,3,4$) and two distances $d_j$ ($j = 1,2$) from the camera, and three laser edge points ${^L}\mathbf{p}_k$ ($k = 1,2,3$) from the LRF.

\begin{figure}[tb]
	\centering
	\includegraphics[width=0.75\columnwidth]{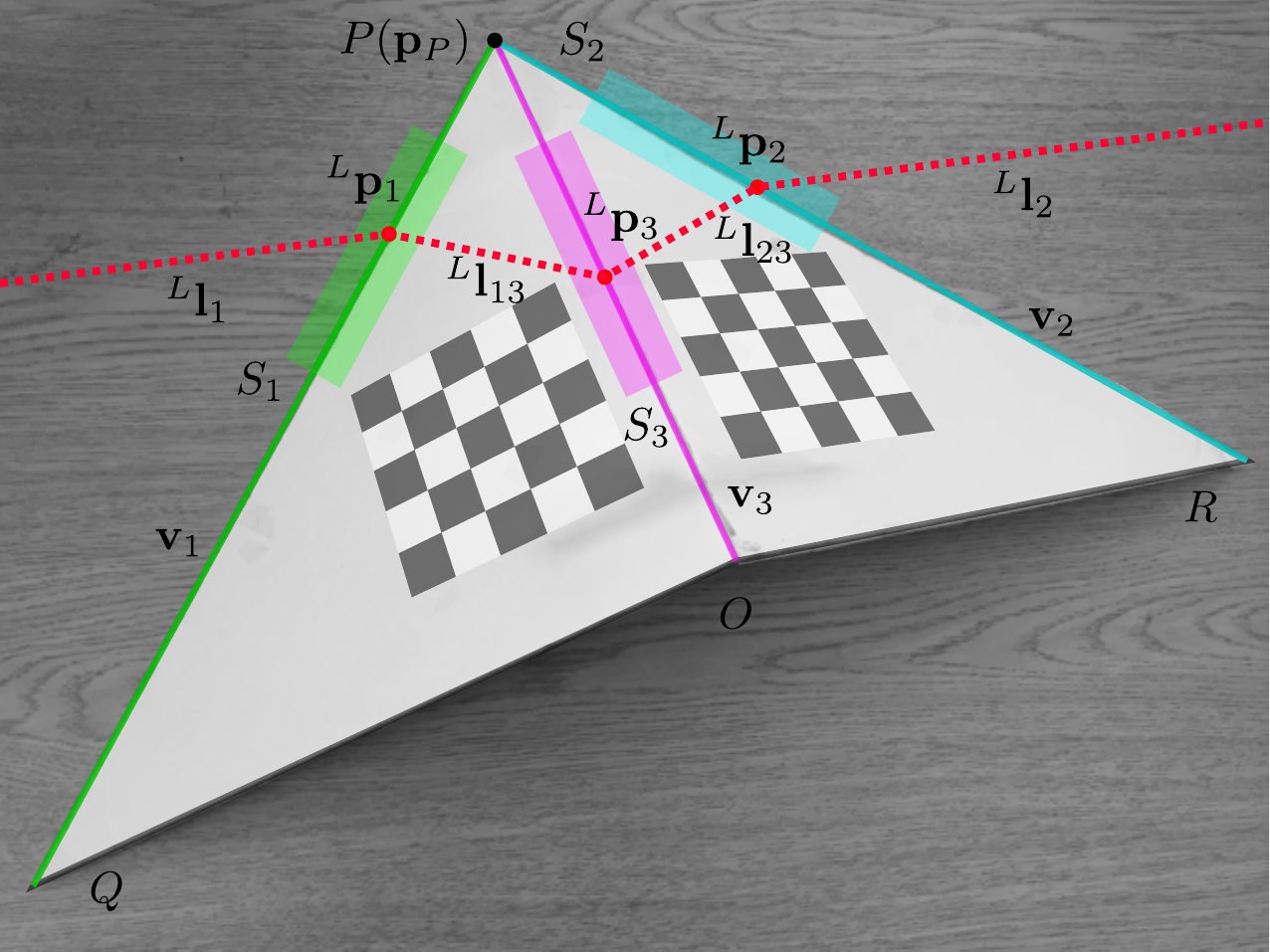}
	\caption{A calibration target built in a convex V-shape is put on the table. Accurate data extraction from both the LRF and the camera is performed by line intersection and optimized line detection.}
	\label{fig:dataProcess}
	\vspace*{-2mm}
\end{figure}

\subsection{Data from the LRF}
Our calibration can be built in a convex shape as shown in Fig.~\ref{fig:dataProcess}. We can put the calibration target on a planar object (such as table, wall and the ground) such that ${^L}\mathbf{p}_1$, ${^L}\mathbf{p}_2$ and ${^L}\mathbf{p}_3$ are accurately determined by the intersection between line segments. Specifically, the laser intersection with the objects are first segmented into four parts ${^L}\mathbf{l}_{1}$, ${^L}\mathbf{l}_{13}$, ${^L}\mathbf{l}_{23}$, and ${^L}\mathbf{l}_{2}$ (e.g. using the Ramer-Douglas-Peucker algorithm (RDP) algorithm~\cite{Douglas}). We then fit a line to each segment using the total least squares method~\cite{Golub}. ${^L}\mathbf{p}_1$ is thus obtained by the intersection of lines ${^L}\mathbf{l}_1$ and ${^L}\mathbf{l}_{13}$. Similarly, ${^L}\mathbf{p}_2$ and ${^L}\mathbf{p}_3$ are respectively determined by the other two pairs $\{ {^L}\mathbf{l}_{2}, {^L}\mathbf{l}_{23} \}$ and $\{ {^L}\mathbf{l}_{13}, {^L}\mathbf{l}_{23} \}$.

\begin{figure}[t]
	\centering
	\includegraphics[width=0.95\columnwidth]{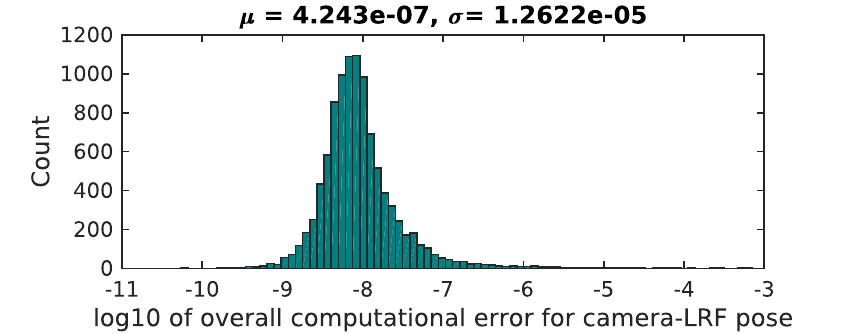}
	\caption{The histogram of computational errors for computed rotation and translation in $10^4$ random and noise-free Monte-Carlo trials.}
	\label{fig:computationalErr}
	\vspace*{-1mm}
\end{figure}

\subsection{Data from the Camera}
Using our calibration target with two checkerboards, the camera calibration is first performed by the MATLAB Toolbox~\cite{calibToolbox}. Two normal vectors ${^C}\mathbf{n}_3$ and ${^C}\mathbf{n}_4$ of planes $T_3$ and $T_4$ can be obtained directly from the calibration toolbox. Each checkerboard is on the plane $z_T = 0$ in its own world frame.
For example, we have the rotation ${_{T_3}^C}\mathbf{R}$ and translation $^C\mathbf{t}_{T_3}$ from the calibration result, which represent the orientation and position of the plane $T_3$ w.r.t. the camera. ${^C}\mathbf{n}_3$ is just minus the 3rd column of ${_{T_3}^C}\mathbf{R}$, and the distance $d_1$ from the camera to $T_3$ is $d_1 = {^C}\mathbf{n}_3^{\top} \cdot ^C\mathbf{t}_{T_3}$. Similarly, ${^C}\mathbf{n}_4$ and $d_2$ can be also calculated.

From the image (See Fig.~\ref{fig:dataProcess}), the unit line directions of $\overline{PQ}$, $\overline{PR}$ and $\overline{PO}$ are projected as $\mathbf{v}_1$, $\mathbf{v}_2$ and as $\mathbf{v}_3$, and the projection point of the vertex $P$ is $\mathbf{p}_P$. 
As stated in~\cite{Ruben}, we perform a weighted optimization initialized by the LSD~\cite{Gioi} line detector to accurately estimate the set $\{ \mathbf{p}_P, \mathbf{v}_1, \mathbf{v}_2, \mathbf{v}_3 \}$. Specifically, only the pixels within a rectangular region $S_i$ are considered for fitting $\mathbf{v}_i$. Let $\boldsymbol{\eta}_i$ be the normal of $\mathbf{v}_i$. Hence, the optimization problem is expressed as
\begin{equation}
\argmin_{\mathbf{p}_P, \mathbf{v}_1, \mathbf{v}_2, \mathbf{v}_3} \sum_{i=1}^3 \sum_{j=1}^{N_i} G_i^j \cdot ( ( \mathbf{p}_i^j - \mathbf{p}_P ) \cdot \boldsymbol{\eta}_i )^2 ,
\end{equation}
where each region $S_i$ has the number $N_i$ of valid pixels $\mathbf{p}_i^j$ whose gradient magnitudes $G_i^j$ as their weights are above a threshold. 
Given the intrinsic matrix $\mathbf{K}$, the normal vector ${^C}\mathbf{n}_i$ ($i = 1, 2$) is obtained by ${^C}\mathbf{n}_i = \frac{\mathbf{K}^{-1}\mathbf{v}_i}{\Vert \mathbf{K}^{-1}\mathbf{v}_i \Vert}$.

\subsection{Snapshot Selection}
The solution from a single snapshot constrains three laser points ${^L}\mathbf{p}_k$ ($k = 1,2,3$). 
In the presence of noise, it should also guarantee that laser points ${^L}\mathbf{p}_{13}^i$ from the line ${^L}\mathbf{l}_{13}$ and ${^L}\mathbf{p}_{23}^j$ from ${^L}\mathbf{l}_{23}$ must respectively lie on $T_3$ and $T_4$. 
We determine a snapshot as ill-conditioned if the average squared distance of laser points to their corresponding planes is larger than a threshold $\epsilon^2$. 
From multiple observations, it can guide us to select well-conditioned snapshots for further accuracy. Given the solution ${_L^C}\hat{\mathbf{R}}$ and ${^C}\hat{\mathbf{t}}{_L}$, we will keep this snapshot if it satisfies
\begin{equation} \label{selectSnapshot}
\scalebox{0.65}
{$
\sum\limits_{i=1}^{N_{13}} \dfrac{( {^C}\mathbf{n}_3^{\top} ( {_L^C}\hat{\mathbf{R}} \cdot {^L}\mathbf{p}_{13}^i + {^C}\hat{\mathbf{t}}{_L} ) - d_1 )^2}{2N_{13}} + \sum\limits_{j=1}^{N_{23}} \dfrac{( {^C}\mathbf{n}_4^{\top} ( {_L^C}\hat{\mathbf{R}} \cdot {^L}\mathbf{p}_{23}^j + {^C}\hat{\mathbf{t}}{_L} ) - d_2 )^2}{2N_{23}} \le \epsilon^2
$} ,
\end{equation}
where ${^L}\mathbf{l}_{13}$ and ${^L}\mathbf{l}_{23}$ are treated equally, and have $N_{13}$ and $N_{23}$ points, respectively. The average squared distance of each line is first calculated and then half weighted as the left two terms of summation in Eq.~(\ref{selectSnapshot}).

\section{Synthetic Experiments} \label{experimentSR}

\begin{figure*}
	\centering
	\begin{minipage}[t]{0.99\columnwidth}
		\includegraphics[width=0.92\columnwidth]{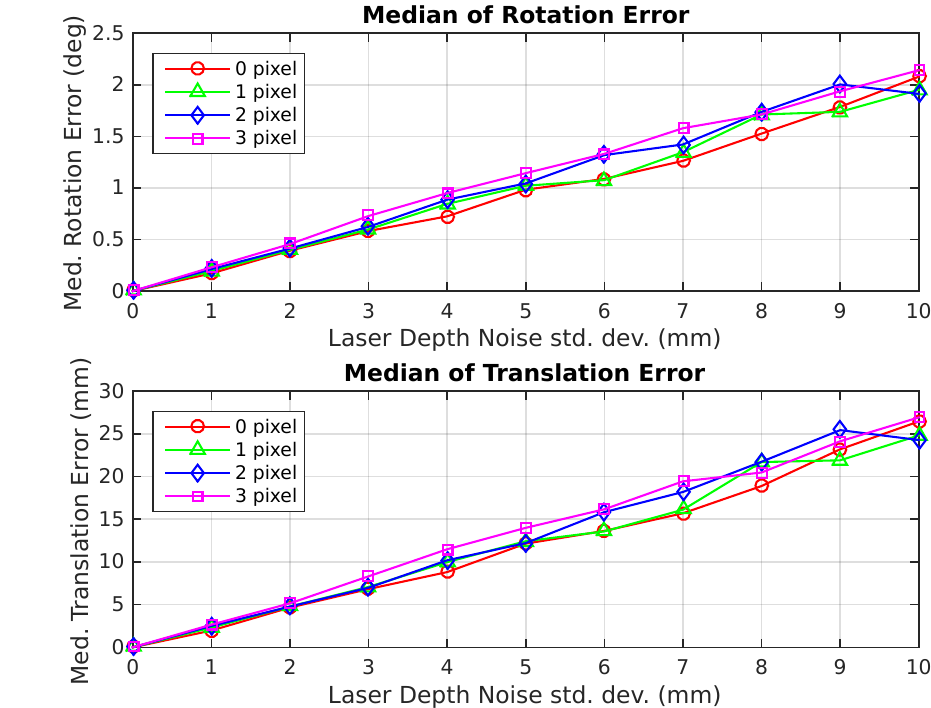}
		\caption{Errors in estimated rotation and translation as a function of the laser noise with different levels of the pixel noise. Each point represents the median error for 1000 Monte-Carlo trials. Each line corresponds to a different level of the pixel noise versus the laser noise changed by their own STDs.} \label{fig:AngleDistErr}
	\end{minipage} \quad
	\centering
	\begin{minipage}[t]{0.99\columnwidth}
		\includegraphics[width=0.92\columnwidth]{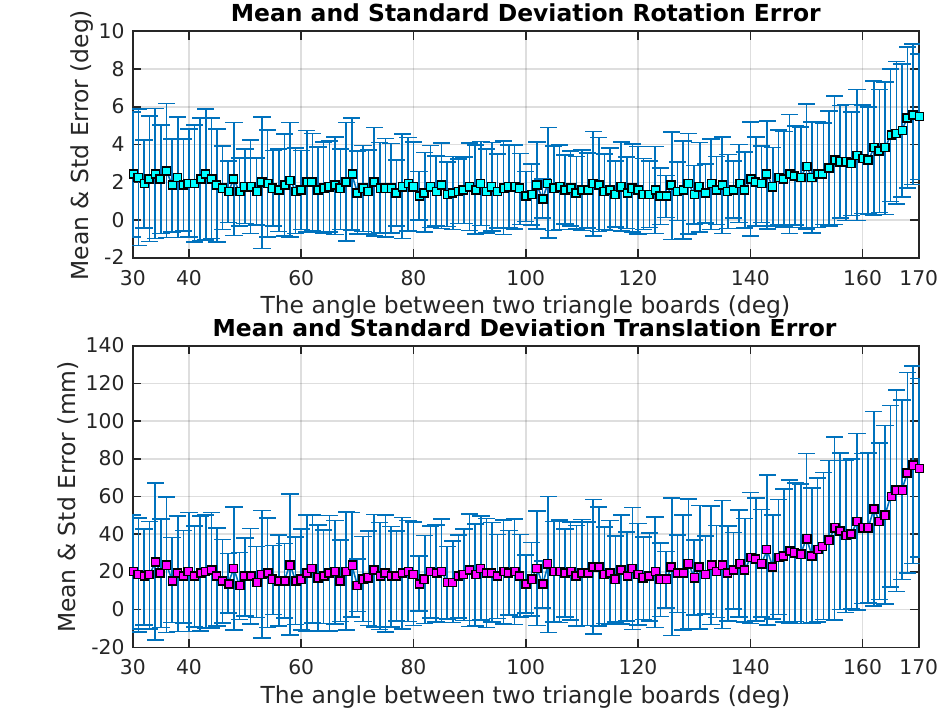}
		\caption{Means and STDs of the errors for estimated rotation and translation versus the angle between two triangle boards of the calibration target. Each point represents the mean error with its STD from 1000 Monte-Carlo trials. The errors gently vary as long as the angle between boards is obtuse.} \label{fig:angleBoards}
	\end{minipage}
	\vspace*{-2mm}
\end{figure*}

\begin{figure*}
	\centering
	\begin{minipage}[t]{0.99\columnwidth}
		\centering
		\includegraphics[width=0.89\columnwidth]{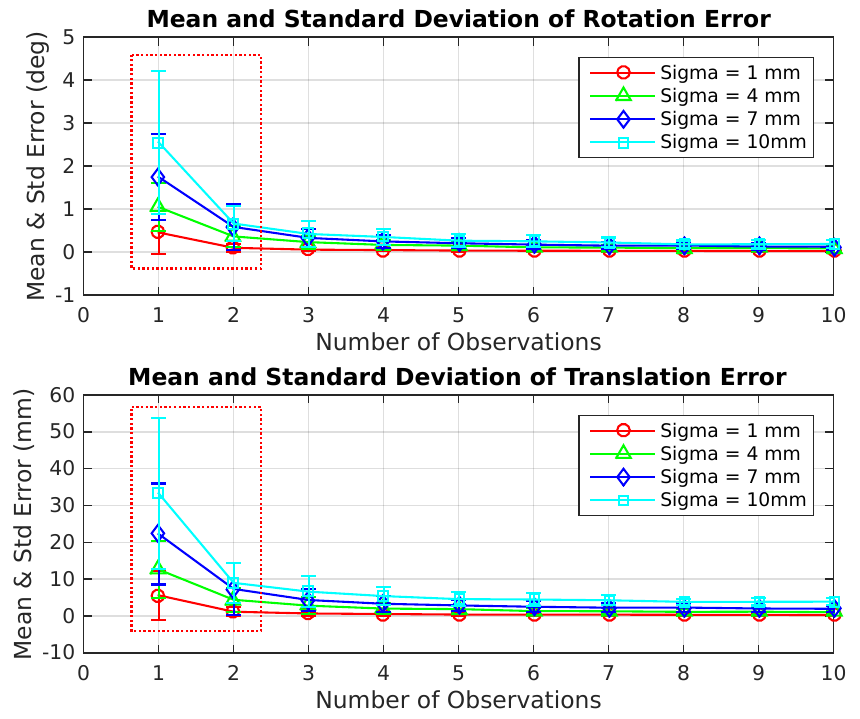}
		\caption{Means and STDs of the errors for estimated rotation and translation versus the number of observations of the calibration target in 1000 Monte-Carlo trials with the laser noise changed by its STD.} \label{fig:NoiseReduction}
	\end{minipage} \quad
	\centering
	\begin{minipage}[t]{0.99\columnwidth}
		\centering
		\includegraphics[width=0.89\columnwidth]{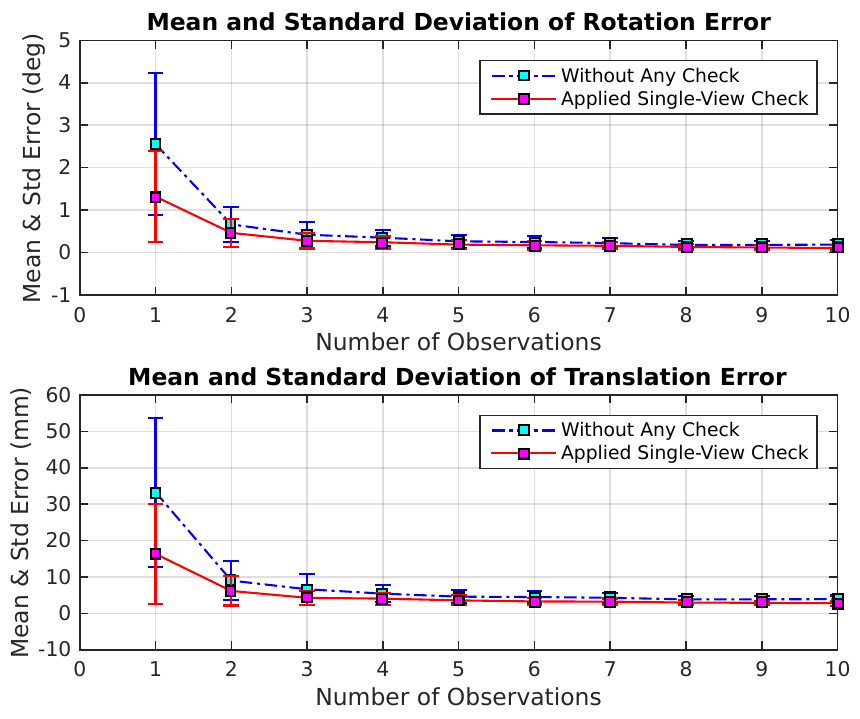}
		\caption{The comparision of means and STDs of the errors for estimated rotation and translation between single-view-check and without-any-check versus the number of observations of the calibration target in 1000 Monte-Carlo trials.} \label{fig:singleViewCheck}
	\end{minipage}
	\vspace*{-2mm}
\end{figure*}

For the simulation setting, the obtuse angle between two triangle boards of the calibration target is set to $150^{\circ}$. Then, we uniformly and randomly generate roll, pitch and yaw in the range of $\pm 45^{\circ}$ for the orientation of the LRF w.r.t. the camera, and three components of the position from 5 to 30 cm. For each instance of the ground truth, we randomly generate the orientation and position of the calibration target within the range of $\pm 45^{\circ}$ and 50 to 150 cm. In the extreme case, the position angle between the LRF-to-target direction and the camera-to-target direction is more than $33^{\circ}$. It includes the scenario in proportion that the camera-to-LRF distance is large, as long as the target is visible from two sensors.

\subsection{Numerical Stability}
We performed $10^4$ Monte-Carlo trials to validate the numerical stability of our solution in the noise-free case. For each trial, only one snapshot of the calibration target is needed. The error metric $e = \Vert [\mathbf{R}_{gt}|\mathbf{t}_{gt}] - [\hat{\mathbf{R}}|\hat{\mathbf{t}}] \Vert_{\mathcal{F}}$ is the Frobenius norm ($\mathcal{F}$) of the difference between the ground truth ($gt$) and our solution. The histogram of errors is shown in Fig.~\ref{fig:computationalErr}. The computational error varies but the accuracy is still high (around $10^{-8}$). It demonstrates that our method correctly solves the LRF-camera calibration problem, which further validates the sufficiency of the constraints from a single snapshot.

\begin{figure*}[t]
	\centering
	\includegraphics[width=1.0\textwidth]{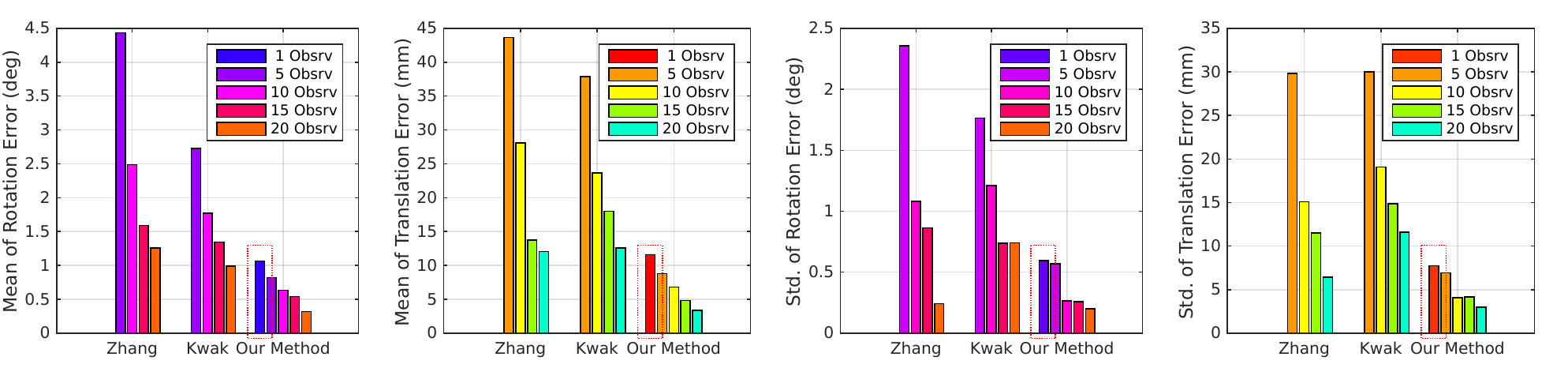}
	\caption{Comparisons with the method of Zhang and Pless~\protect\cite{Zhang} and the method of Kwak et al.~\protect\cite{Kwak}. 1st and 2nd: Means of estimated rotation and translation errors for all three methods as the number of observations of the calibration target increases; 3rd and 4th: STDs of estimated rotation and translation errors for all three methods as the number of observations increases.}
	\label{fig:angleMean}
	\vspace*{-1mm}
\end{figure*}

\subsection{Sensitivity to Data Noise}
This simulation tests the sensitivity of our solution to the noise in feature data. Two sources of error are taken into account: the laser depth uncertainty along the beam direction and the pixel uncertainty in line detection. Based on the practical setting, we respectively set the standard deviations (STDs) of laser noise and pixel error as $\sigma_1 = 10$ mm and $\sigma_2 = 3$ px in the $640\times 480$ image. A factor $k_{\sigma}$ varies from 0 to 1 to combine the noise information as $k_{\sigma}\sigma_i$ ($i = 1, 2$)~\cite{Ruben} from both sensors in one plot. 1000 Monte-Carlo trials are performed, each of which needs only one snapshot. The metrics for angular error (the chordal distance~\cite{Hartley} of rotation) and distance error (translation) are respectively $e_{\theta} = 2\arcsin ( \frac{1}{2\sqrt{2}} \Vert \hat{\mathbf{R}}-\mathbf{R}_{gt} \Vert_{\mathcal{F}} )$ and $e_d = \Vert \hat{\mathbf{t}}-\mathbf{t}_{gt} \Vert_2$. Fig.~\ref{fig:AngleDistErr} demonstrates that as the noise level increases, our solution is robust to the image noise but has a greater sensitivity to the laser depth noise.

\subsection{Sensitivity to Angle between Boards}
The sensitivity of our solution to the angle between two triangle boards of the calibration target is tested. We set $\sigma_1 = 10$ mm and $\sigma_2 = 3$ px. The angle varies from $30^{\circ}$ to $170^{\circ}$. Fig.~\ref{fig:angleBoards} shows the means and STDs for both angular and distance error. Our solution is not sensitive to the angle between boards but still has the smallest errors around $135^{\circ}$.

\subsection{Noise Reduction}
We report a simulation designed for testing the noise reduction of our solution when using multiple observations. Here, $\sigma_1$ varies from 1 mm to 10 mm with $\sigma_2$ set to 3 px. At each noise level, the means and STDs are calculated for both $e_{\theta}$ and $e_d$ from 1000 Monte-Carlo trials. From Fig.~\ref{fig:NoiseReduction}, we observe that as number of observations increases, the means and STDs of both errors asymptotically decrease. Moreover, with a small number of snapshots, our solution can achieve a highly accurate initial value for further optimization. Specifically, $e_{\theta}$ and $e_d$ are respectively around $0.5^{\circ}$ and 5 mm for only 5 snapshots.

\subsection{Snapshot Selection}
With snapshot selection, additional accuracy is achieved (See Fig.~\ref{fig:singleViewCheck}) in the presence of noise $\sigma_1 = 10$ mm and $\sigma_2 = 3$ px. The distance threshold $\epsilon$ is set to 5 mm. The errors $e_{\theta}$ and $e_d$ are further decreased by nearly $50\%$ compared with the solution without any single-snapshot check.
Specifically, $e_{\theta}$ and $e_d$ are respectively around $0.25^{\circ}$ and 2 mm for only 5 snapshots.
Thus, our method can guarantee a high accuracy of the solution when using multiple snapshots.

\section{Real Experiments} \label{realExperiment}
To further validate our method, we perform real experiments in comparison with other two existing methods~\cite{Zhang} and~\cite{Kwak}.

A LRF Hokuyo URG-04LX is rigidly mounted on a stereo rig which consists of a pair of Point Grey Chameleon CMLN-13S2C cameras (See Fig.~\ref{fig:cameralaserRig}). Sensors are synchronized based on time stamps. The LRF is set to $180^\circ$ horizontal field of view, with an angular resolution of $0.36^\circ$ and a line scanning frequency of 10 Hz. Its scanning accuracy is $\pm 1$ cm within a range from 2 cm to 100 cm, and has $1\%$ error for a range from 100 cm to 400 cm. The cameras have a resolution of $640\times 480$ pixels, and are pre-calibrated based on~\cite{ZZhang}. The images prior to data processing are warped to get rid off the radial and tangent distortions.

\subsection{Comparison with Existing Methods}
We compare our method to two state-of-the-art algorithms~\cite{Zhang} and~\cite{Kwak} using the ground truth of the stereo-rig baseline. Specifically, let $\mathbf{T} = \left[\begin{smallmatrix} \mathbf{R} &\mathbf{t} \\ 0 0 0 &1 \end{smallmatrix} \right]$ represent the transformation (rotation and translation) between two frames. For each method, the LRF is first calibrated w.r.t both left and right cameras to obtain $_{L}^{C_l}\mathbf{T}$ and $_{L}^{C_r}\mathbf{T}$. We then compute the relative pose (baseline) between stereo cameras and compare it with the the ground truth $_{C_l}^{C_r}\mathbf{T}$ calibrated from the toolbox~\cite{calibToolbox}. Hence, the error matrix is $\mathbf{T}_e = _{C_l}^{C_r}\mathbf{T} \cdot _{L}^{C_l}\mathbf{T} \cdot ( _{L}^{C_r}\mathbf{T} )^{-1}$, where $\mathbf{R}_e$ and $\Vert \mathbf{t}_e \Vert_2$ are respectively compared with the identity and zero.

The stereo cameras are calibrated for 10 times where each time 20 image pairs are randomly chosen from 40 pairs. With the rotation represented by the Rodrigues' formula, the means of the rotation angle and the baseline distance are respectively $0.0137^\circ$ and 96.0511 mm (the translation is [$-96.0505$, $-0.3035$, $-0.1326$]$^{\top}$ in mm). Because of their low STDs $0.0018^\circ$ and 0.2742 mm (within $0.01^\circ$ and 1 mm), we treat their means as the ground truth. The distances between the LRF and the stereo cameras are approximately 100 mm and 150 mm. 30 best observations of each method are obtained using a RANSAC framework (5 snapshots are randomly chosen from a total 50). We randomly select subsets of 1, 5, 10, 15 and 20 observations to perform the calibration between the LRF and the stereo rig. The calibration is performed 10 times for each random subset.

Fig.~\ref{fig:angleMean} shows that our method has the smallest errors of both rotation and translation as the number of observations increases. Specifically, the mean errors from 20 observations are respectively $0.3^\circ$ and $3.4$ mm almost three times lower than the method of Zhang and Pless ($1.3^\circ$ and $12.0$ mm) and the method of Kwak et al. ($1.0^\circ$ and $12.6$ mm). Moreover, our method can obtain a reasonable result even using only one snapshot, which is \emph{not possible} for the other two methods. In other words, our method can achieve an accuracy at the same level but using the smallest number of observations.

\subsection{Real Scene Validation}
\begin{figure}[t]
	\centering
	\includegraphics[width=0.93\columnwidth]{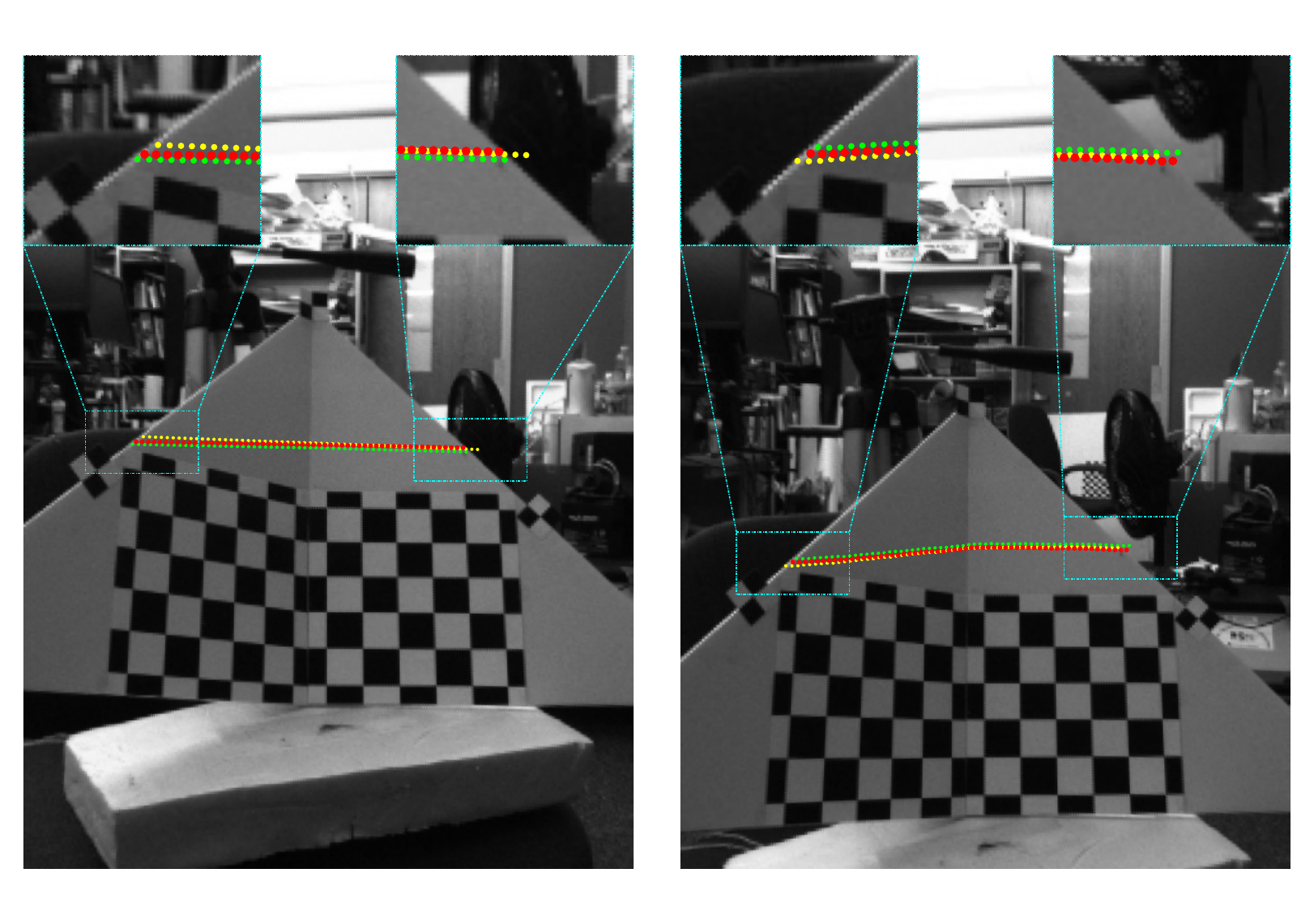}
	\caption{The projection of laser points scanned on the calibration target onto the images from both left and right cameras using three different methods. Cyan boxes show the close-up of laser point projection. Methods~\protect\cite{Zhang} and~\protect\cite{Kwak} are green and yellow colored, respectively. Our approach is red colored.}
	\label{fig:backProjection}
	\vspace*{-2mm}
\end{figure}

This experiment from the real scene tests the calibration results between LRF to both left and right cameras, respectively, obtained by two existing methods~{\cite{Zhang}}~{\cite{Kwak}} and our method for comparison (See Fig.~{\ref{fig:backProjection}}). We generate a dataset of 30 input observations for each method using their own calibration targets. The calibration results are obtained from 20 observations randomly chosen from 30 in total of each method. We then respectively test them using the stereo images from new observations of our calibration target which are not involved in the calibration process.
From the comparison, we observe that the laser scanning lines for our method more reasonably match the calibration target (board boundaries) from both left and right cameras. Thus, it validates the correctness of our calibration results of each LRF-camera pair.

\section{Conclusion} \label{concl}
In this paper, we presented a novel method for calibrating the extrinsic parameters of a system of a camera and a 2D laser rangefinder.
In contrast to existing methods, our coplanarity constraints for feature data suffice to unambiguously determine the relative pose between these two sensors even from a single observation. A series of experiments verified that the number of observations can be drastically reduced for an accurate result. Our solution technique was also extended to the case of multiple observations to reduce noise for further refinement.
We are working on releasing our code~\cite{Code}.

\appendix
\section{Proof of Lemma 1} \label{NonIndependence}
From Fig.~\ref{fig:calibTarget}, ${^C}\mathbf{n}_i$ is the normal vector of plane $T_i$ for $i = {1,2,3,4}$ and we claim that these normal vectors in any cardinality three subset of $\{{^C}\mathbf{n}_1, {^C}\mathbf{n}_2, {^C}\mathbf{n}_3, {^C}\mathbf{n}_4\}$ are linearly independent. It is obvious that there are totally four subsets: I. ${^C}\mathbf{n}_1$, ${^C}\mathbf{n}_2$ and ${^C}\mathbf{n}_3$; II. ${^C}\mathbf{n}_1$, ${^C}\mathbf{n}_2$ and ${^C}\mathbf{n}_4$; III. ${^C}\mathbf{n}_1$, ${^C}\mathbf{n}_3$ and ${^C}\mathbf{n}_4$; IV. ${^C}\mathbf{n}_2$, ${^C}\mathbf{n}_3$ and ${^C}\mathbf{n}_4$. We will prove separately for each subset and show that subsets I and II are symmetric arguments, and subsets III and IV are also symmetric arguments.

According to the geometry setting in Fig.~\ref{fig:calibTarget}, we notice that for each subset three different planes have a common intersection point $P$, which means there is no parallelism between them. We let $\mathbf{l}_{12}$, $\mathbf{l}_{13}$, $\mathbf{l}_{23}$ and $\mathbf{l}_{34}$ respectively denote the directional unit vectors of the lines $\overline{PC}$, $\overline{PQ}$, $\overline{PR}$ and $\overline{PO}$ w.r.t. the camera frame. 

So let us first prove the claim for subset I: ${^C}\mathbf{n}_1$, ${^C}\mathbf{n}_2$ and ${^C}\mathbf{n}_3$. We assume that these three normal vectors are linearly dependent, which means there exists three nonzero coefficients $\alpha$, $\beta$ and $\gamma$ such that $\alpha {^C}\mathbf{n}_1 + \beta {^C}\mathbf{n}_2 + \gamma {^C}\mathbf{n}_3 = 0$,
otherwise two of three planes would have parallelism (e.g. $\alpha = 0$ such that $\beta {^C}\mathbf{n}_2 = -\gamma {^C}\mathbf{n}_3$) or one plane of them reduces to nonexistence (e.g. $\alpha = \beta = 0$ such that $\gamma {^C}\mathbf{n}_3 = 0$). Thus, ${^C}\mathbf{n}_3$ can be represented as the combination of ${^C}\mathbf{n}_1$ and ${^C}\mathbf{n}_2$. Since the intersecting line of $T_1$ and $T_2$ is $\overline{PC}$, we have the following
\begin{equation}
\begin{cases}
\mathbf{l}_{12}^{\top} \cdot {^C}\mathbf{n}_1 = 0 \\
\mathbf{l}_{12}^{\top} \cdot {^C}\mathbf{n}_2 = 0 
\end{cases} \quad \Rightarrow \quad \mathbf{l}_{12}^{\top} \cdot {^C}\mathbf{n}_3 = 0 .
\end{equation}
It means that the line $\overline{PC}$ is on the plane $T_3$ given the fact that they share a common point $P$. It is a contradiction unless camera center $C$ is also on the plane $T_3$, which is a useless case since camera cannot capture the checkerboard on $T_3$. Thus, the normal vectors ${^C}\mathbf{n}_1$, ${^C}\mathbf{n}_2$ and ${^C}\mathbf{n}_3$ are linearly independent. It is similar for subset II that we would have a contradiction that the line $\overline{PC}$ is on the plane $T_4$ based on $\mathbf{l}_{12}^{\top} \cdot {^C}\mathbf{n}_4 = 0$ if ${^C}\mathbf{n}_1$, ${^C}\mathbf{n}_2$ and ${^C}\mathbf{n}_4$ are linear dependent. So we can conclude that these vectors in both subset I and II are linearly independent.

Next let us focus on the claim for subset III: ${^C}\mathbf{n}_1$, ${^C}\mathbf{n}_3$ and ${^C}\mathbf{n}_4$. We assume that these three normal vectors are linearly dependent, which means there exists three nonzero coefficients $\alpha$, $\beta$ and $\gamma$ as explained in subset I such that $\alpha {^C}\mathbf{n}_1 + \beta {^C}\mathbf{n}_3 + \gamma {^C}\mathbf{n}_4 = 0$.
Thus, ${^C}\mathbf{n}_1$ can be represented as the combination of ${^C}\mathbf{n}_3$ and ${^C}\mathbf{n}_4$. Since the intersecting line of $T_3$ and $T_4$ is $\overline{PO}$, we have the following
\begin{equation}
\begin{cases}
\mathbf{l}_{34}^{\top} \cdot {^C}\mathbf{n}_3 = 0 \\
\mathbf{l}_{34}^{\top} \cdot {^C}\mathbf{n}_4 = 0 
\end{cases} \quad \Rightarrow \quad \mathbf{l}_{34}^{\top} \cdot {^C}\mathbf{n}_1 = 0 .
\end{equation}
It means that the line $\overline{PO}$ is on the plane $T_1$ given the fact that they share a common point $P$. It is a contradiction unless the corner $O$ of the calibration target is also on the plane $T_1$, which is a useless case since camera cannot capture the checkerboard on $T_3$. Thus, the normal vectors ${^C}\mathbf{n}_1$, ${^C}\mathbf{n}_3$ and ${^C}\mathbf{n}_4$ are linearly independent. It is similar for subset IV that we would have a contradiction that the line $\overline{PO}$ is on the plane $T_2$ based on $\mathbf{l}_{34}^{\top} \cdot {^C}\mathbf{n}_2 = 0$ if ${^C}\mathbf{n}_2$, ${^C}\mathbf{n}_3$ and ${^C}\mathbf{n}_4$ are linear dependent. So we can conclude that these vectors in both subset III and IV are also linearly independent.

Above all, it is proved that three normal vectors in each subset from I, II, III and IV are linearly independent. \QEDA

\section{Proof of Lemma 2} \label{NonSingular}
From Appendix~\ref{NonIndependence}, we know that any three of normal vectors ${^C}\mathbf{n}_1$, ${^C}\mathbf{n}_2$, ${^C}\mathbf{n}_3$ and ${^C}\mathbf{n}_4$ can span the whole 3D space. Based on the definitions in~(\ref{reDefinition}), the matrix in~(\ref{translation}) is 
\begin{equation}
\scalebox{0.75}
{$
\mathcal{N}_o = \sum_{i=1}^N {^C}\bar{\mathbf{n}}_i {^C}\bar{\mathbf{n}}_i^{\top} = {^C}\mathbf{n}_1{^C}\mathbf{n}_1^{\top}+{^C}\mathbf{n}_2{^C}\mathbf{n}_2^{\top}+2{^C}\mathbf{n}_3{^C}
\mathbf{n}_3^{\top}+2{^C}\mathbf{n}_4{^C}\mathbf{n}_4^{\top}
$} ,
\end{equation}
which is a symmetric matrix. We now show that this matrix is non-singular.

As is known, the eigenvalues of a positive definite matrix are all positive~\cite{PDM}.
Further, we know that a positive definite matrix is always invertible~\cite{PDM}. 
From the properties above, we just need to prove that $\mathcal{N}_o$ is positive definite. Let $\mathbf{v}\neq \mathbf{0}$ be an arbitrary non-zero vector. Then we calculate the quadratic form
\begin{equation}
\scalebox{0.8}
{$
\mathbf{v}^{\top}\mathcal{N}_o\mathbf{v} = \left({^C}\mathbf{n}_1^{\top}\mathbf{v}\right)^2+\left({^C}\mathbf{n}_2^{\top}\mathbf{v}\right)^2+2\left(
{^C}\mathbf{n}_3^{\top}\mathbf{v}\right)^2+2\left({^C}\mathbf{n}_4^{\top}\mathbf{v}\right)^2 \ge 0
$} .
\end{equation}
We assume that $\mathbf{v}^{\top}\mathcal{N}_o\mathbf{v} = 0$, which means ${^C}\mathbf{n}_i^{\top}\mathbf{v} = 0$ for $i = \{ 1,2,3,4 \}$. However, since any three of these four normal vectors are linearly independent, we get a contradiction that, for example, $[{^C}\mathbf{n}_1, {^C}\mathbf{n}_2, {^C}\mathbf{n}_3]^{\top}\mathbf{v} = \mathbf{0}$ if and only if $\mathbf{v} = \mathbf{0}$. Thus, we can conclude that
\begin{equation}
\scalebox{0.8}
{$
\mathbf{v}^{\top}\mathcal{N}_o\mathbf{v} = \left({^C}\mathbf{n}_1^{\top}\mathbf{v}\right)^2+\left({^C}\mathbf{n}_2^{\top}\mathbf{v}\right)^2+2\left(
{^C}\mathbf{n}_3^{\top}\mathbf{v}\right)^2+2\left({^C}\mathbf{n}_4^{\top}\mathbf{v}\right)^2 > 0
$} .
\end{equation}
Thus, matrix $\mathcal{N}_o$ is positive definite and always invertible. \QEDA

%
%
%

%
%
%


\begin{biography}{Wenbo Dong} 
is a PhD student in the Department of Computer Science and Engineering at the University of Minnesota, Twin Cities. He obtained his MS degree (2017) in Computer Science at the University of Minnesota. He obtained his BE (2012) and ME (2014) degrees in Electrical Engineering at Harbin Institute of Technology, Harbin, China.
His research interests include robotics, 3D computer vision and their applications in precision agriculture.
\end{biography}

\begin{biography}{Volkan Isler}
is a Professor in the Department of Computer Science and Engineering at the University of Minnesota. He obtained his MSE (2000) and PhD (2004) degrees in Computer and Information Science from the University of Pennsylvania. 
He obtained his BS degree (1999) in Computer Engineering from Bogazici University, Istanbul, Turkey.
In 2008, he received the National Science Foundation's Young Investigator Award (CAREER). From 2009 to 2015, he chaired IEEE Society of Robotics and Automation's Technical Committee on Networked Robots. He also served as an Associate Editor for IEEE Transactions on Robotics and IEEE Transactions on Automation Science and Engineering. He is currently an Editor for the ICRA Editorial Board. His research interests are primarily in robotics, computer vision, sensor networks and geometric algorithms, and their applications in agriculture and environmental monitoring.
\end{biography}

\end{document}